\documentclass{article}

    \usepackage[preprint]{neurips_2025}

\usepackage[utf8]{inputenc} % allow utf-8 input
\usepackage[T1]{fontenc}    % use 8-bit T1 fonts
\usepackage{hyperref}       % hyperlinks
\usepackage{url}            % simple URL typesetting
\usepackage{booktabs}       % professional-quality tables
\usepackage{amsfonts}       % blackboard math symbols
\usepackage{nicefrac}       % compact symbols for 1/2, etc.
\usepackage{microtype}      % microtypography
\usepackage{xcolor}         % colors

\usepackage{graphicx}
\let\cite\citep

\newcommand{\parencite}{\citep}
\newcommand{\textcite}{\citet}
\def\ie{\textit{i.e.}}
\def\eg{\textit{e.g.}}

\newcommand{\beginSupplementaryMaterials}{
    \clearpage
    \begin{center}
        \textbf{\LARGE Supplementary Materials}\\
        \vspace{0.1cm}
    \end{center}
    \vspace{-.5cm}
    \noindent\makebox[\linewidth]{\rule{\textwidth}{0.4pt}}
}

\newcommand{\EnvState}{s^E}
\newcommand{\EnvAction}{a^E}
\newcommand{\EnvReward}{r^E}

\newcommand{\MonState}{s^M}
\newcommand{\MonAction}{a^M}
\newcommand{\MonReward}{r^M}

\newcommand{\ProxyReward}{\hat{r}^E}
\newcommand{\undefined}{\bot}
\newcommand{\monitor}{\mathcal{M}}

\newcommand{\argmax}{arg\,max}

\usepackage{subcaption}
\usepackage{adjustbox}
\usepackage{nicefrac}
\usepackage{siunitx}
\usepackage{mathtools}
\usepackage{textgreek}
\usepackage{booktabs}
\usepackage{wrapfig}

\title{Generalization in Monitored Markov Decision Processes (Mon-MDPs)}

\author{
    Montaser Mohammedalamen\\
        University of Alberta\\
        Edmonton, Alberta, Canada\\
        \texttt{mohmmeda@ualberta.ca}\\
        \And
    Michael Bowling\\
    University of Alberta; Amii\\
    Edmonton, Alberta, Canada\\
        \texttt{mbowling@ualberta.ca}
}
\begin{document}

\maketitle

\begin{abstract}
Reinforcement learning (RL) typically models the interaction between the agent and environment as a Markov decision process (MDP), where the rewards that guide the agent's behavior are always observable. However, in many real-world scenarios, rewards are not always observable, which can be modeled as a monitored Markov decision process (Mon-MDP). Prior work on Mon-MDPs have been limited to simple, tabular cases, restricting their applicability to real-world problems. This work explores Mon-MDPs using function approximation (FA) and investigates the challenges involved. We show that combining function approximation with a learned reward model enables agents to generalize from monitored states with observable rewards, to unmonitored environment states with unobservable rewards. Therefore, we demonstrate that such generalization with a reward model achieves near-optimal policies in environments formally defined as unsolvable. However, we identify a critical limitation of such function approximation, where agents incorrectly extrapolate rewards due to \emph{overgeneralization}, resulting in undesirable behaviors. To mitigate overgeneralization, we propose a cautious police optimization method leveraging reward uncertainty. This work serves as a step towards bridging this gap between Mon-MDP theory and real-world applications.
\end{abstract}

\section{Introduction}\label{sec:introduction}
Reinforcement learning (RL) has emerged as a powerful framework for solving complex decision-making problems, achieving superhuman performance in a wide range of games, such as Atari games~\citep{bellemare2013arcade, mnih2015human}, Go~\citep{silver2016mastering}, StarCraft~\citep{vinyals2019grandmaster}, Gran Turismo~\citep{wurman2022outracing}, and poker~\citep{moravvcik2017deepstack}. This success has fueled interest in applying RL beyond simulated environments to real-world applications, where it has demonstrated remarkable capabilities. Notable examples include optimizing data center cooling~\citep{evans2023optimizing}, controlling water treatment~\citep{janjua2024gvfs}, navigating stratospheric balloons~\citep{bellemare2020autonomous}, adapting prosthetic devices~\citep{pilarski2011online}, and enabling robots to play table tennis semi-competitively~\citep{d2024achieving}.
% and even discovering efficient algorithms for matrix multiplication~\citep{fawzi2022discovering}.

A common feature across these examples is that they modeled the interaction between the agent and environment as the Markov decision process (MDP) framework, where an agent observes a state from an environment, takes an action, and receives an immediate reward as feedback. The agent aims to maximize the expected (discounted) cumulative reward. However, many real-world applications involve environments where rewards are not always observable. For example, consider an autonomous plant-watering robot responsible for maintaining plant hydration in a home. This robot learns from feedback, such as a positive reward from a human when it waters a plant or data from a monitoring system that measures soil moisture.
However, continuous human feedback is often impractical due to time constraints, and installing sensors for each plant may not be cost-effective. Despite this, the agent should continue watering plants appropriately, even when rewards are unobservable, \eg, when homeowners are away.
% as shown in the experiments section.
To address the challenge of unobservable rewards, the monitored-MDPs (Mon-MDPs) framework was introduced~\cite{parisi2024first}. Mon-MDPs model the interaction between the agent and environment when rewards are not always observable. However, prior work on Mon-MDPs~\citep{parisi2024first,parisi2024last,aliriza2005} have been limited to simple, tabular environments with a finite, trackable number of states and actions. This limitation highlights a gap between the current state of Mon-MDP research and its original motivation of modeling complex real-world applications where traditional MDPs cannot be applied.
% Many real-world applications are non-tabular or involve high-dimensional states (\eg, images).

This work takes an important first step towards bridging this gap by exploring Mon-MDPs in non-tabular settings, using function approximation (FA), and investigating the associated challenges with generalization.
% Generalization can be used to infer estimated rewards when the reward is unobservable.
% \MET{Above, we just argued about high-dimensional states. Are we only focusing on non-tabular in this paper?}
We also demonstrate that FA can lead to \emph{overgeneralization}, where rewards are incorrectly extrapolated from monitored to unmonitored states, potentially leading to undesirable or unsafe behaviors. To address overgeneralization, we adopt an algorithm that leverages reward uncertainty and robust optimization, showing that agents can \emph{learn} to act cautiously in unmonitored states, mitigating unsafe behaviors.
We view extending Mon-MDPs to continuous action spaces as a promising direction for future research.
% Extending Mon-MDPs to continuous action spaces is left for future work.
This paper's key contributions are summarized as follows:

% This paper aims to investigate the challenges associated with Mon-MDPs in non-tabular settings and to understand the capabilities and limitations of function approximation within the Mon-MDP framework. Therefore, our key contributions are summarized as follows:
\begin{enumerate}
    \item \label{contribution:1} We demonstrate that incorporating FA and a learned reward model leads to near-optimal policies in some Mon-MDPs. In contrast, approaches that treat Mon-MDPs as traditional MDPs (\eg, ignoring unobservable rewards or assuming they are zeros) can yield sub-optimal policies. This replicates observations of \cite{parisi2024first} in non-tabular settings.
    \item \label{contribution:2} We empirically show that combining the FA and a reward model enables agents to generalize from monitored states with observable rewards, to unmonitored states with unobservable rewards. This generalization can allow agents to achieve near-optimal performance in environments formally defined as unsolvable Mon-MDPs in the tabular setting.
    \item \label{contribution:3} However, we also show that FA may lead agents to incorrectly extrapolate rewards, resulting in undesirable behaviors. To address \emph{overgeneralization}, we adapt an algorithm that leverages reward uncertainty and robust policy optimization. Our approach enables agents to \emph{learn} to act cautiously when rewards are unobservable, mitigating the risk of overgeneralization.
\end{enumerate}

This paper is structured as follows: We begin in Section~\ref{sec:background} by reviewing background for MDPs, Monitored-MDPs, robust policy optimization, and FA.
Section~\ref{sec:method} introduces our approach for extending the reward model to non-tabular settings using FA.
Experiments and results are presented in Section~\ref{sec:experiments}, demonstrating the effectiveness of our approach.
Section~\ref{sec:conclusion} outlines the method's limitations and potential directions for future research.
Finally, Section~\ref{sec:broaderImpact} reflects on the broader impact.

\section{Background}\label{sec:background}
This section provides background on MDPs as a framework for modeling agent-environment interactions and introduces the Mon-MDPs framework, which extends MDPs to settings where rewards are not always observable. Additionally, we discuss the related work on generalization in RL using FA. Finally, we explore robust policy optimization techniques that address uncertainty challenges.

\subsection{Markov decision processes}\label{subsubsec:mdp}
RL traditionally frames the interaction between the agent and environment as an MDP. An MDP is a mathematical framework for sequential decision-making, defined by a tuple $(\mathcal{S}, \mathcal{A}, \mathcal{R}, \mathcal{P}, \gamma)$. At each step $t$, the agent receives a state $s_t \in \mathcal{S}$ from the environment, takes an action $a_t \in \mathcal{A}$, and receives a single scalar reward $r_t \in \mathcal{R}$ sampled from the reward function $r_t \sim \mathcal{R}(s_t, a_t)$. The agent then transitions to the next state $s_{t+1} \in \mathcal{S}$ according to a Markovian state transition probability distribution $s_{t+1} \sim \mathcal{P}(s_t, a_t)$. Therefore, the agent acts in the environment according to a policy, which probabilistically maps a state to an action with a probability $\pi(a_t | s_t)$, as shown in Figure~\ref{fig:mdp_diagram}.
% The reward is discounted according to a discount factor $\gamma \in [0, 1)$ to balance between immediate and future rewards; if $\gamma = 0$, the agent will maximize the immediate reward only. However, as $\gamma$ increases, the agent will maximize a longer horizon of future rewards.
To evaluate the policy, a state value function is defined as the expected discounted return,
$V_{\pi}(s) = \mathbb{E}_{\pi} \left[\sum_{t = 1}^{\infty} \gamma^{t-1} r_t \text{$|$} s_t=s\right]$. Similarly, a state-action value (Q-value) is the value of being in state $s$ and taking action $a$ under policy $\pi$, denoted as,
$Q_{\pi}(s, a) = \mathbb{E}_{\pi} \left[\sum_{t = 1}^{\infty} \gamma^{t-1} r_t \text{$|$} s_t=s, a_t=a\right]$.
According to the Bellman optimality equation~\cite{bellman1957dynamic}, there is a unique optimal value function for each MDP $Q_{*}(s, a) = \sum_{s_{t+1}} \mathcal{P}(s_{t+1}|s_t, a_t) \left[ \mathcal{R}(r_t | s_t, a_t) + \gamma \max_{a_{t+1}} Q_{*}\left(s_{t+1}, a_{t+1}\right) \right]$, and, there is at least one optimal policy $\pi_{*}(s_t) = \argmax_{a_t} Q_{t}(s_t, a_t)$.
% $\left[ \mathcal{R}(r_{t} | s_{t}, a_{t}) + \gamma \sum_{s_{t+1}} \mathcal{P}(s_{t+1}|s_t, a_t) V_{*}(s_{t+1}) \right]$. Therefore, the agent's goal is to find the optimal policy or the optimal state value function.

\subsection{Monitored Markov decision processes}\label{subsubsec:mon_mdp}
% \MET{its weird that the AAMAS paper, labeled 2024b came out before the NeurIPS paper, labeled 2024a. maybe we can change the ordering in the bibtex?} 

\begin{figure*}[tb]
  \centering
  \vspace{-0.8cm}
  \begin{adjustbox}{valign=t,minipage=0.25\textwidth}
         \includegraphics[width=\textwidth]{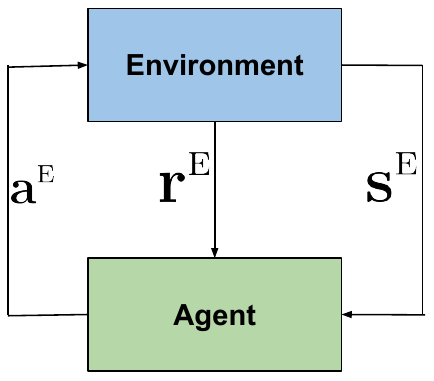}
         \subcaption{MDP Diagram.}
        \label{fig:mdp_diagram}
    \end{adjustbox}
  % \hfill
  \hspace{0.8cm}
  \begin{adjustbox}{valign=t,minipage=0.37\textwidth}
         \vspace{-0.35cm}
         \includegraphics[width=\textwidth]{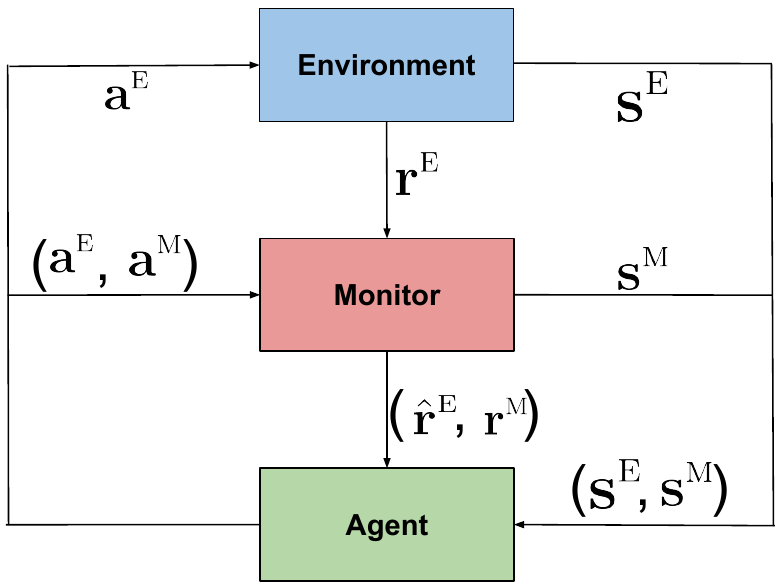}
         \subcaption{Monitored-MDP Diagram.}
         \label{fig:mon_mdp_diagram}
    \end{adjustbox}
        \caption{MDP and Monitored-MDP Diagrams.}
        \label{fig:mdp_vs_mon_diagram}
        % \vspace{-0.5cm}
\end{figure*}

% \begin{wrapfigure}{R}{0.5\textwidth}
%     \begin{adjustbox}{valign=t,minipage=0.5\textwidth}
%         \centering
%         \vspace{-1.8cm}
%         \includegraphics[width=0.65\textwidth]{fig/mon_mdp/mdp.pdf}
%         \subcaption{\footnotesize{MDP}}
%         \label{fig:mdp_diagram}
%     \end{adjustbox}
%     \hfill
%     \begin{adjustbox}{valign=t,minipage=0.5\textwidth}
%         \centering
%          \includegraphics[width=0.75\textwidth]{fig/mon_mdp/mon_mdp.pdf}
%          \subcaption{\footnotesize{Monitored-MDP}}
%          \label{fig:mon_mdp_diagram}
%     \end{adjustbox}
%         \caption{MDP \& Monitored-MDP Diagrams.}
%         \label{fig:mdp_vs_mon_diagram}
%         \vspace{-0.5cm}
% \end{wrapfigure}

Many real-world applications cannot be modeled as traditional MDPs because rewards are not always observable by the agent. The Mon-MDP framework~\cite{parisi2024first} addresses this limitation by defining the agent-environment interaction as involving both an environment MDP and a separate monitor MDP that controls when the reward is observed (see Figure~\ref{fig:mon_mdp_diagram}).  Consider the tuple $(\mathcal{S}^{E}, \mathcal{A}^{E}, \mathcal{P}^{E}, \mathcal{R}^{E}, \mathcal{M}, \mathcal{S}^{M}, \mathcal{A}^{M}, \mathcal{P}^{M}, \mathcal{R}^{M}, \gamma)$, where $(\mathcal{S}^{E}, \mathcal{A}^{E}, \mathcal{P}^{E}, \mathcal{R}^{E}, \gamma)$ represents the environment MDP, whereas $(\mathcal{S}^{M}, \mathcal{A}^{M}, \mathcal{P}^{M}, \mathcal{R}^{M}, \gamma)$ represents the monitor MDP. 

The transition in Mon-MDPs can be divided into a transition to the next environment state, depending on the current environment state and environment action $\EnvState_{t+1} \sim \mathcal{P}^{E}(\EnvState_{t}, \EnvAction_{t})$, and a transition to the next monitor state, which depends on both the environment and monitor state and action $\MonState_{t+1} \sim \mathcal{P}^{M}(\EnvState_{t}, \MonState_{t}, \EnvAction_{t}, \MonAction_{t})$. Unlike traditional MDPs, the agent does not directly observe the reward from the environment $r^E_t \sim \mathcal{R}^E(s^E_t, a^E_t)$, but instead observes the \emph{proxy reward}, which is generated by a Markovian monitor function $\mathcal{M}$. The proxy denoted as $\ProxyReward_t \sim \monitor(\EnvReward_t, \MonState_t, \MonAction_t)$, can be either observable $\ProxyReward_t \in \mathbb{R}$ or unobservable $\ProxyReward_t = \undefined$.
Therefore, at any timestep $t$, the agent receives the environment and monitor states $(\EnvState_t, \MonState_t)$, takes actions in both the environment and monitor $(\EnvAction_t, \MonAction_t)$, and receives a tuple of a proxy reward and a monitoring reward $(\ProxyReward_t, \MonReward_t)$. However, the agent's goal is to maximize $\mathbb{E}_{\pi} \left[\sum_{t = 1}^{\infty} \gamma^{t-1} (\EnvReward_t + \MonReward_t)\right]$, i.e., the sum of monitoring reward and the actual (possibly unobserved) environment reward.
In this sense, MDPs are considered a special case of Mon-MDPs, where the reward is always observable $\ProxyReward=\EnvReward$, the monitoring reward $\MonReward=0$, and there is a single monitoring state and action $|\mathcal{A}^{M}| = |\mathcal{S}^{M}| = 1$.

While Mon-MDPs may appear conceptually similar to partially observable MDPs (POMDPs)~\citep{aastrom1965optimal}, the key distinction lies in the nature of partial observability: in POMDPs, the environment state is partially observable, whereas in Mon-MDPs, it is the reward that could be unobservable.
% that is partially observable.

Mon-MDPs can be categorized into two distinct types based on the existence of a discernible optimal policy~\citep{parisi2024first}: i) \emph{solvable Mon-MDPs}, where at least one optimal policy exists for all indistinguishable~\cite{binmore2007game} reward functions; ii) \emph{unsolvable Mon-MDPs}, where there is no policy that is optimal for all indistinguishable reward functions (\ie, the agent can never know if its policy is optimal). An extreme case of unsolvable Mon-MDPs is \emph{hopeless Mon-MDPs}, where all policies are optimal for some indistinguishable reward function (\ie, any policy for the agent could be optimal). An example of a hopeless Mon-MDP is a scenario where rewards are never observable.
% leaving the agent unable to infer which actions contribute to achieving the desired objective.

To address the challenge of reward observability in Mon-MDPs,~\cite{parisi2024first} proposed learning a tabular reward model that estimates the environment reward and incorporates the estimated values into the value function update. While this approach is theoretically proven to converge to an optimal policy for solvable Mon-MDPs in tabular settings,
the tabular reward model requires states and actions to be finite, with the reward model and value functions explicitly represented as tables.
% its effectiveness is contingent upon several restrictive assumptions that significantly limit its applicability to real-world scenarios. Specifically:
% the reward model’s convergence guarantees hold only under the following stringent conditions:
Specifically, the tabular reward model convergence guarantees hold only if:
i) the environment is tabular with a finite or countably infinite~\cite{halmos1960countable}
% (a set can be placed in a one-to-one correspondence with the set of natural numbers $\mathbb{N}$)
number of states and actions; ii) each joint state pair $(\EnvState, \MonState)$ can be visited infinitely often given infinite exploration; iii) the agent can observe the environment reward for each environment state-action pair with some nonzero probability; and iv) the proxy reward is truthful, meaning $\ProxyReward_t=\EnvReward$ or $\ProxyReward_t=\bot$ $\forall t$.

These assumptions reveal significant limitations in existing work on reward models for Mon-MDPs, as many real-world applications involve non-tabular or infinite state spaces. Ensuring sufficient visitation for all joint state pairs becomes infeasible, complicating the exploration and exploitation trade-off. Nonetheless, requiring the agent to observe the environment reward for every state-action pair further restricts the reward model's applicability. To overcome these limitations and relax these strong assumptions, it is necessary to extend the reward model using function approximation, thereby enabling learning in complex and non-tabular environments.
% \MET{I don't understand the second clause - why is observing rewards accurately infeasible now?}
% , thereby undermining the reward model's ability to generalize.

% \begin{figure*}[tb]
%   \centering
%   \begin{adjustbox}{valign=t,minipage=0.335\textwidth}
%          \includegraphics[width=\textwidth]{fig/mon_mdp/mdp.pdf}
%          \subcaption{\footnotesize{MDP Diagram: the agent executes actions in the environment and receives the state and the reward.}}
%          \label{fig:mdp_diagram}
%     \end{adjustbox}
%     \hspace*{0.55cm}
%     \begin{adjustbox}{valign=t,minipage=0.6\textwidth}
%          \includegraphics[width=0.65\textwidth]{fig/mon_mdp/mon_mdp.pdf}
%          \subcaption{\footnotesize{Monitored-MDP Diagram: the monitor takes the environment reward along with the joint state and action and outputs the proxy and monitor rewards.}}
%          \label{fig:mon_mdp_diagram}
%     \end{adjustbox}
%         \caption{MDP Vs Monitored-MDP Diagrams.}
%         \label{fig:mdp_vs_mon_diagram}
% \end{figure*}

\subsection{Function approximation}\label{subsubsec:fa}
Many real-world problems involve high-dimensional, possibly continuous states or actions.
% For instance, a $19\times19$ Go game board has approximately $3^{361}$ states.
To address this complexity, function approximation aims to automatically learn to identify similarities, differences, and relationships between states and actions.
% In supervised learning, FA maps between features (inputs) and targets (outputs), where targets can be a vector of real numbers in the case of regression, \eg, expected discounted return from a state following each action.
% or classes (labels) in the case of classification.
A function approximator is denoted as $y = f(x, \theta)$, where $y$ is the target, $x$ is the input features, and $\theta$ are the function's parameters $f$. The ultimate goal of FA is not simply to memorize examples from the training data
% fit inputs to targets,
but to capture the underlying data-generating process, thereby enabling generalization to unseen examples.

FA is particularly beneficial in RL when dealing with high-dimensional state spaces (\eg, images) or large or continuous action spaces, enabling agents to generalize and scale to complex environments where tabular methods are impractical.
% \MET{This distinction between states and actions is weird to me. Don't we care about the state space being large and/or continuous, and care about the action space being large and/or continuous?}
Therefore, FA can be applied in several key areas: First, approximate value functions, where $\hat{V}(s, \theta)\approx V(s)$ or $\hat{Q}(s, a, \theta)\approx Q(s, a)$ and $\theta \in \mathbb{R}^{d}$, with fewer parameters than states $d \ll |\mathcal{S}|$, this approximation can be achieved using either linear methods~\citep{boyan1994generalization,gordon1995stable,sutton1988learning,papavassiliou1999convergence} or non-linear function approximation~\citep{ernst2005tree,lange2012batch,mnih2013playing}. Second, parameterize the policy, presenting the policy as a distribution $a \sim \pi(s, \theta)$~\citep{konda1999actor,Sutton2000,schulman15trpo,schulman2017proximal,haarnoja2018soft}.
Third, learn environment models, such as the reward function $\hat{\mathcal{R}}(s, a, \theta)$ and transition function $\hat{\mathcal{P}}(s, a, \theta)$~\citep{kuvayev1996model,sutton2008dyna,deisenroth2011PILCO,levine2013,silver2016mastering,Hafner2020Dream}.

% Therefore, FA can be used to find an approximate solution for: i) value functions: $\hat{V}(s, \theta)\approx V(s)$ or $\hat{Q}(s, a, \theta)\approx Q(s, a)$, where $\theta \in \mathbb{R}^{d}$ and the number of parameters is smaller than the number of states $d \ll |\mathcal{S}|$ using linear function approximation~\citep{sutton1988learning,boyan1994generalization,gordon1995stable,papavassiliou1999convergence} and non-linear function approximation~\citep{,mnih2013playing}; ii) parameterize the policy: $\pi(a| s, \theta)$~\citep{Sutton2000,schulman15trpo, schulman2017proximal, konda1999actor, haarnoja2018soft};
% % Although, we will use a parameterize policy in this work;
% and iii) learning a mode of the reward function $\hat{r}(s, a, \theta)$ or a model of the transition function $\hat{\mathcal{P}}(s, a, \theta)$~\citep{kuvayev1996model,deisenroth2011PILCO,levine2013,silver2016mastering,sutton2008dyna,Hafner2020Dream}.
% \MET{Is $d$ defined?}
% \MET{Section 2.1 added $\pi$ to V and Q. I'd make it the same here.}
% Generalization is the premise behind using function approximation.

\subsection{Robust policy optimization}\label{subsubsec:robuts_policy_optimization}
Despite FA's ability to handle high-dimensional data and generalize, it may produce incorrect predictions, particularly for novel, out-of-distribution examples. This uncertainty is known as \emph{epistemic uncertainty}, which arises from insufficient knowledge. For example, the agent may be uncertain about the probability distribution of a dice roller rather than the outcome of the next roll.

In RL various approaches address uncertainty through robust policy optimization, enabling agents to learn policies that perform well under uncertainty, adversarial conditions, or distributional shifts~\citep{russel2020robust,jiang2021Monotonic,Kuang_Lu_Wang_Zhou_Li_Li_2022}.
% within the reward model. To introduce a risk-averse approach within Mon-MDPs we adopt the 
To specifically address epistemic uncertainty, the ``learning to be cautious'' framework was initially proposed for a risk-averse approach within MDPs~\citep{mohammedalamen2021learning}. This framework combines an ensemble of neural networks to quantify epistemic uncertainty~\citep{tibshirani1996comparison,heskes1997practical,lakshminarayanan2017simple,lu2017ensemble,pearce2018bayesian,osband2019deepExplorationRvf,lee2021sunrise} and applies conditional value at risk (CVaR) optimization, utilizing the $k$-of-$N$ counterfactual regret minimization (CFR)~\citep{kofn}.
This method enables agents to learn to act cautiously in uncertain environments. In this approach, $k$ and $N$ are integers,  $k > 0$ and $N \ge k$. The $k$-of-$N$ CFR algorithm samples $N$ reward models and updates the policy based on the average performance of the $k$-worst models. The $\nicefrac{k}{n}$ ratio controls the robustness level: $k = N$ corresponds to risk-neutral behavior, while a smaller $k$ with a larger $N$ induces a more risk-averse policy.
% By optimizing for worst-case outcomes, this method enables agents to identify and adopt cautious, robust behaviors in environments characterized by epistemic uncertainty.
Although the ``learning to be cautious'' framework was originally designed for traditional MDPs, we extend it to Mon-MDPs to introduce risk aversion for novel states, where we may never obtain an accurate reward estimate.
% as it assumes that rewards are undefined (unobservable) for novel states.
This extension aligns naturally with the Mon-MDPs setting.

% \MET{Before we dive into experiments, shouldn't we talk about what our contribution is? I was expecting to see a new algorithm or a description of how we modified old algos so that they could use function approximation.} 

\section{Reward model with function approximation}\label{sec:method}
To enable agents to generalize in Mon-MDPs, we extend the previously proposed tabular reward model~\citep{parisi2024first} by incorporating FA.
This extended reward model is denoted as $\hat{\mathcal{R}}(\EnvState, \EnvAction, \theta)$, which takes the environment state $\EnvState$ and the environment action $\EnvAction$ as input. This model is trained using samples when the proxy reward is equal to the environment reward $\ProxyReward=\EnvReward$.
In addition to the reward model, we also train a Q-model, denoted as $\hat{Q_{\pi}}(\EnvState, \MonState, \EnvAction, \MonAction, \theta)$, incorporating both the environment and monitor states as input $(\EnvState, \MonState)$ to estimate the Q-value for each joint environment-monitor action $(\EnvAction, \MonAction)$.

Specifically, both the reward model and Q-model are implemented using a neural network architecture composed of two convolutional layers followed by two fully-connected layers.
The Q-model follows the deep Q-Network (DQN) approach~\citep{mnih2013playing}, which employs a target network and a Q-network, trained using mini-batches from a replay experience buffer.
For exploration, we use $\epsilon$-greedy with $\epsilon$ linearly decayed.
% from $1$ to $0.1$ over $10^4$ timesteps and then remains constant.
% of size $128$
% capable of storing up to $10^6$ samples.
Further details on the experimental setup, model training procedure, computational resources, and hyper-parameters tuning are provided in the Appendix~\ref{app:experiments}.
% \footnote{Our code and models are available anonymously at the following link \url{https://shorturl.at/aQF9V}.}.
However, our implementation differs in two key aspects: i) DQN receives the joint state $(\EnvState, \MonState)$ as input to estimate the Q-value for all joint action pairs $(\EnvAction, \MonAction)$; and ii) instead of using the environment reward, DQN utilizes the predicted reward provided by the reward model. Consequently, both the reward model and Q-network are trained simultaneously.

% Following the original implementation of DQN~\citep{mnih2013playing}, 

In Mon-MDPs, the success of reward model generalization depends critically on the: i) state representation; ii) generalizing ability of the function approximator, and iii) regularity (observability) of the reward in the environment. When these things do not align, we call this phenomenon ``overgeneralization'', as the following experiment section will demonstrate.

In addition to the reward model, we evaluate two alternative methods that treat Mon-MDPs as traditional MDPs: i) $\bot=0$, undefined rewards are replaced with $0$ when updating the Q-network; and ii) ``ignore'', updates the Q-network only with samples where rewards are observable, discarding samples with unobservable rewards.

% For a fair comparison, all algorithms utilize the same DQN architecture as the reward model, and performance is reported as the mean and the $95\%$ confidence interval.
% Furthermore, we tune the following hyper-parameters for all methods individually, including the learning rate for both the reward model and Q-network $\eta^{R}, \eta^{Q}$ and the $\epsilon$-decay rate.

\section{Experiments and Results}\label{sec:experiments}
This section presents a series of experiments in non-tabular settings, evaluating the capabilities and limitations of FA in Mon-MDPs. All experiments are based on the plant-watering robot example introduced as a motivation example in Section~\ref{sec:introduction}.

\begin{figure*}[tb]
  \centering
  \vspace{-0.8cm}
  \begin{adjustbox}{valign=t,minipage=0.22\textwidth}
         \includegraphics[width=\textwidth]{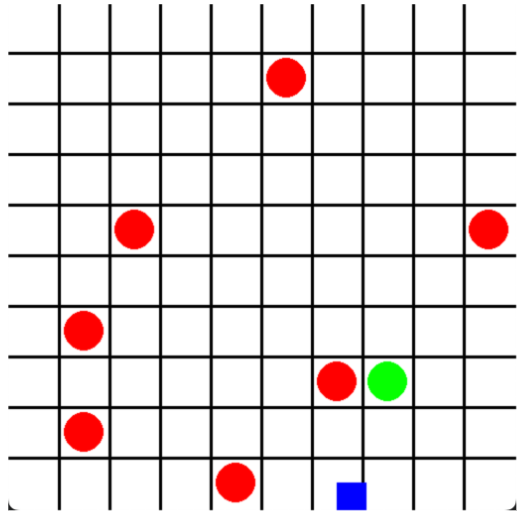}
         \subcaption{\footnotesize{Binary.}}
         \label{fig:binary_env}
    \end{adjustbox}
    \hspace*{0.05cm}
    \begin{adjustbox}{valign=t,minipage=0.25\textwidth}
         \includegraphics[width=\textwidth]{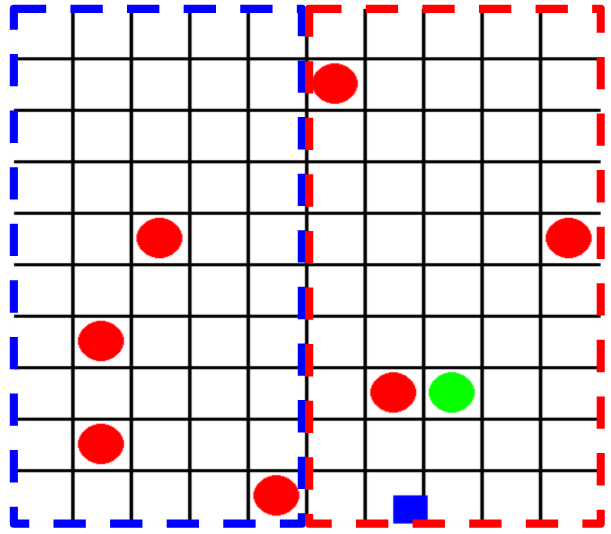}
         \subcaption{\footnotesize{Half-room.}}
         \label{fig:half_room_env}
    \end{adjustbox}
    \hspace*{0.05cm}
    \begin{adjustbox}{valign=t,minipage=0.25\textwidth}
         \includegraphics[width=\textwidth]{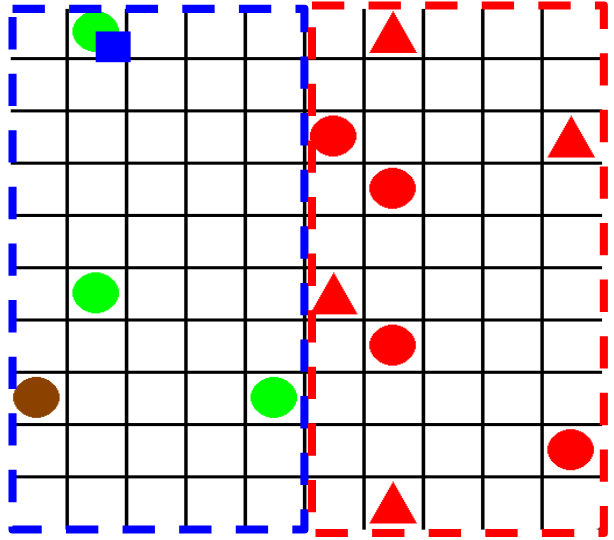}
         \subcaption{\footnotesize{Plants \& cacti.}}
         \label{fig:cacti_room_env}
    \end{adjustbox}
    \hspace*{0.05cm}
    \begin{adjustbox}{valign=t,minipage=0.215\textwidth}
         \includegraphics[width=\textwidth]{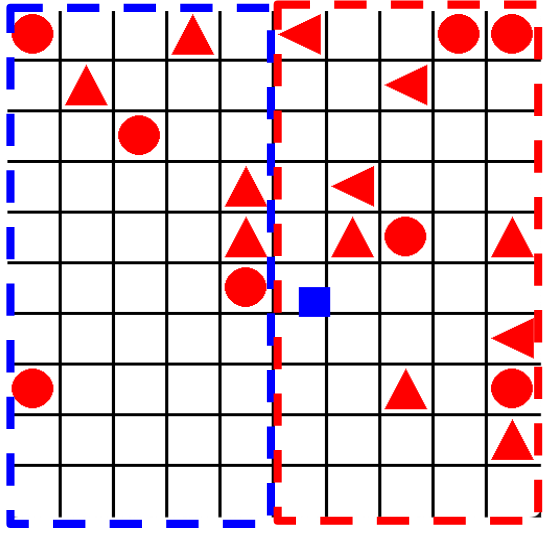}
         \subcaption{\footnotesize{Botanical garden.}}
         \label{fig:3_rooms_env}
    \end{adjustbox}
        \caption{Frames from each environment.}
        \label{fig:envs}
        % \vspace{-0.5cm}
\end{figure*}

The \textbf{plant-watering environment} is a $10\times10$ grid-world, see Figure~\ref{fig:binary_env}.
% environment designed to reflect the motivation example.
In this environment, the agent, represented by a blue square, can perform six discrete actions: $\{\uparrow, \downarrow, \rightarrow, \leftarrow, Water, Stay\}$. The agent aims to maintain the hydration of eight plants, represented as circles, randomly positioned within the grid. Each plant has three levels of dryness $\{$fully dry (red), partially dry (brown), wet (green)$\}$.
The agent observations consist of six distinct feature channels that capture the agent's location, the type of plant (three channels), the plant's dryness level, and the walls. The agent perceives the environment through an egocentric-view with a window size $=11$, where the agent is in the center of the window, resulting in state dimensions of ($6 \times 11 \times 11$). Consequently, this window size does not always provide a complete view of the environment unless the agent is precisely at the center of the grid. This high-dimensional representation captures spatial and contextual information, increasing the complexity of the learning task compared to traditional tabular environments.

At the beginning of each episode, the agent’s location and the plant positions are initialized randomly without replacement to ensure variability across episodes. All plants are initialized fully dry. The agent receives an environment reward of $\EnvReward=+1$ for watering a fully or partially dry plant, and the plant’s dryness level decreases by one unit. However, if the agent waters a wet plant, it will receive a penalty of $\EnvReward=-1$. Moreover, the agent will receive a small penalty of $\EnvReward=-0.2$ for watering an empty cell, effectively spilling water on the floor. Therefore, the agent's goal is to water plants appropriately while avoiding overwatering.
A stochastic drying process causes some plants to become dry. At each timestep, with a probability defined by the dryness rate $=5\%$, a randomly selected plant’s dryness level increases by one unit, reflecting natural dehydration over time. This environment has no termination condition, but the episode is truncated after $100$ timesteps.

\subsection{Binary monitor}\label{subsec:binary_monitor}

The first experiment explores Contribution~\ref{contribution:1}, extending the reward model to non-tabular settings. Specifically, it applies function approximation in the plant-watering environment to enable learning in high-dimensional state spaces.
% such as the plant-watering environment, using function approximation.
% which has previously been shown to converge in tabular settings, .
% Specifically within the plant-watering environment. 
This experiment introduces a binary variant of the plant-watering environment, as shown in Figure~\ref{fig:binary_env}, where the agent has two additional monitoring actions it can choose from along with the six environment actions, $\MonAction \in \{$ask to be monitored, not ask to be monitored$\}$. Therefore, if the agent asks to be monitored, it observes the true environment reward $\ProxyReward=\EnvReward$ and receives a monitoring reward $\MonReward=-0.2$ on that step. On the other hand, choosing not to be monitored results in unobservable rewards $\ProxyReward=\bot$ but avoids the monitoring penalty $\MonReward=0$. Therefore, the optimal policy should be to water all plants while avoiding over-watering, spilling water on the floor, or requesting monitoring to avoid incurring a penalty.
However, the agent will need to request monitoring to be able to learn this policy.

Although the binary environment appears relatively simple, modeling it as a tabular environment requires representing more than $10^{18}$ distinct states. Consequently, storing the corresponding reward model and Q-value table requires an excessively large amount of memory. This challenge emphasizes the critical need for FA, which enables agents to generalize across similar states.

% To overcome the memory challenge and to enable agents to generalize, we extend the previously proposed tabular reward model~\cite{parisi2024first} by incorporating FA. Specifically, we train a reward model composed of two convolutional layers followed by two fully-connected layers. This reward model uses the environment state $\EnvState$ as input and outputs the predicted environment reward for each action. Additionally, we implement a Deep Q-Network (DQN) model, consisting of a target and Q-networks trained using mini-batches sampled from a replay experience buffer, following the approach of~\cite{mnih2013playing}. The primary distinction in our implementation is that DQN receives the joint state $(\EnvState, \MonState)$ as input to estimate the Q-value for all joint action pairs $(\EnvAction, \MonAction)$.

% \TODO{Monta: Move algo before section 3.1}
% \MET{In addition to describing our algorithm, we should probably separate out the discussion of the benchmark algos into their own subsection (rather than being in the Binary Monitor section)}

% Following the original implementation of DQN~\cite{mnih2013playing}.
% Exploration is managed using $\epsilon$-greedy, where $\epsilon$ is linearly decayed from $1$ to $0.1$ over $10^4$ timesteps and then remains constant. During training, the Q-network is updated by sampling batches of size $128$ from a replay buffer capable of storing up to $10^6$ samples. Additional details on the experimental setup, model training procedure, and hyper-parameters can be found in the Appendix.

For a fair comparison, all algorithms utilize the same DQN architecture for the reward model, $\bot=0$, and ``ignore'', and performance is reported as the mean and the $95\%$ confidence interval.
Furthermore, we tune the hyper-parameters for all methods individually, including the learning rate for both the reward model and Q-network $\eta^{R}, \eta^{Q}$ and the $\epsilon$-decay rate (see Appendix~\ref{app:experiments}).

Figure~\ref{fig:binary_train_curve} illustrates the training curve in the binary monitor environment; the reward model outperforms $\bot=0$ and ``ignore'' approaches, with a negligible variation across seeds.
Additionally, Figure~\ref{fig:binary_frequency} shows the frequency of the monitoring action for each policy during training. The reward model rapidly learns to stop requesting monitoring, whereas $\bot=0$ continuously requests monitoring at all times. Interestingly, the ``ignore'' approach selectively requests monitoring only when watering plants, as it only learns from samples where rewards are observable. These results demonstrate that the reward model empirically converges to the optimal policy in the binary environment. Furthermore, other methods that treat Mon-MDPs as traditional MDPs converge to sub-optimal policies. Specifically, $\bot=0$ converges to ask continuously to be monitored, and ``ignore'' learns to ask to be monitored only when watering plants. This highlights the reward model’s ability to generalize effectively in non-tabular environments, addressing Contribution~\ref{contribution:1}.

\begin{figure*}[tb]
  \centering
  % \vspace{-0.8cm}
  \begin{adjustbox}{valign=t,minipage=0.65\textwidth}
         \includegraphics[width=\textwidth]{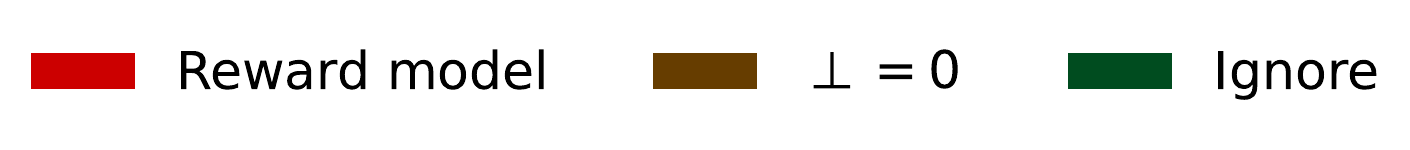}
    \end{adjustbox}
  \hfill
  \begin{adjustbox}{valign=t,minipage=0.42\textwidth}
         \vspace{-0.35cm}
         \includegraphics[width=\textwidth]{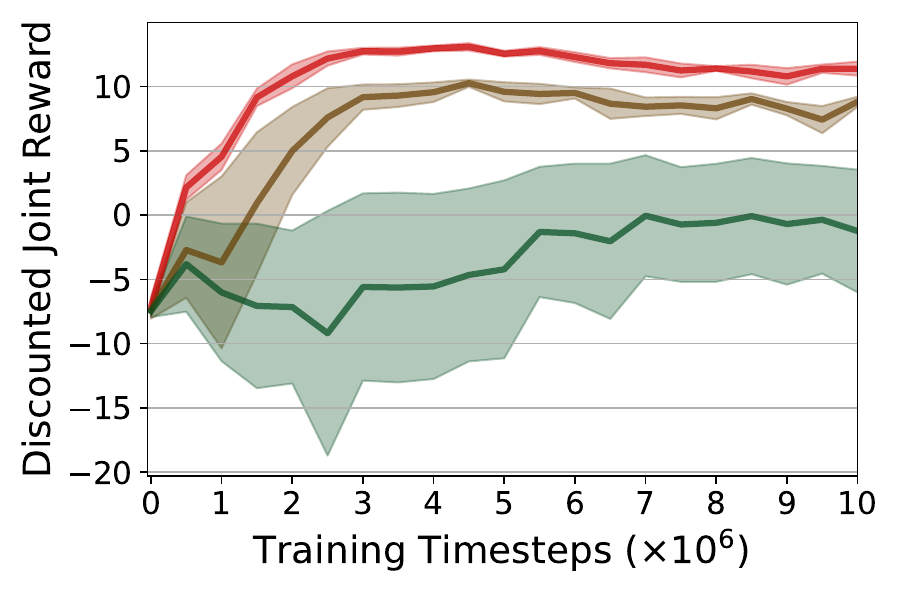}
         \subcaption{\footnotesize{Training curve.}}
         \label{fig:binary_train_curve}
    \end{adjustbox}
    \begin{adjustbox}{valign=t,minipage=0.42\textwidth}
         \vspace{-0.35cm}
         \includegraphics[width=\textwidth]{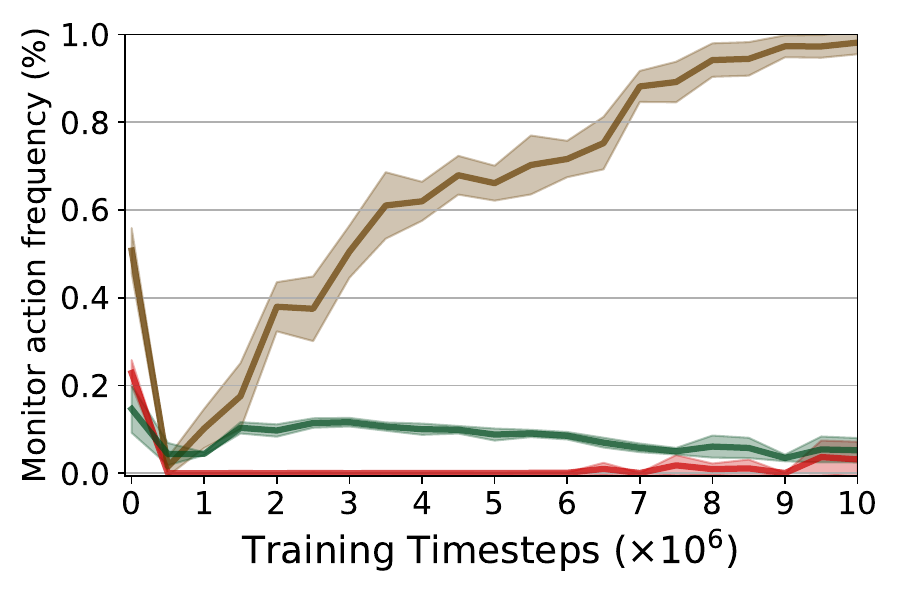}
         \subcaption{\footnotesize{Frequency of monitor action during training.}}
         \label{fig:binary_frequency}
    \end{adjustbox}
        \caption{Reward model, ``ignore'', and $\bot=0$ performance in the binary environment. The bold line represents the mean over $10$ seeds; the shaded area is a $95\%$ interval confidence.}
        \label{fig:binary_train_frequency}
        \vspace{-0.5cm}
\end{figure*}

\subsection{Half-room environment}\label{subsec:half_room_monitor}
The half-room experiment investigates Contribution~\ref{contribution:2}, evaluating whether an agent can generalize from monitored states with observable rewards to unmonitored states with unobservable rewards. To replicate this scenario, we adopt the plant-watering environment, introducing the half-room environment. In this setup, the agent does not have a monitoring action as part of its action space. Monitoring is automatically applied only in the left-half of the room, as illustrated in Figure~\ref{fig:half_room_env}. In the monitored region (Zone $1$), the monitor state is active (``on'') $\Rightarrow\MonState=1$, and the proxy reward equals the true environment reward $\ProxyReward=\EnvReward$ and $\MonReward=0$.
However, when the agent moves to the right-half of the room (Zone $2$), the monitor state is inactive (``off'') $\Rightarrow\MonState=0$, in this region, the proxy reward is undefined $\ProxyReward=\bot$, and the monitor reward is zero $\MonReward=0$.
The half-room environment mimics a scenario in which monitoring (\eg, from human observation or monitoring sensors) is only available in a specific part of the environment. In this setup, the agent must learn to generalize its behavior to the unmonitored region with unobservable rewards. Consequently, the optimal policy in the half-room environment involves properly watering plants, regardless of location.
% This experiment could also serve as an example of the train-then-deploy scenario, where the agent receives a reward during the training in the left-half, then evaluated without a reward during the deployment phase in the right-half.

\begin{figure*}[tb]
  \centering
  \vspace{-0.8cm}
  \begin{adjustbox}{valign=t,minipage=0.47\textwidth}
         \vspace*{0.4cm}
         \hspace*{0.8cm}
         \includegraphics[width=0.8\textwidth]{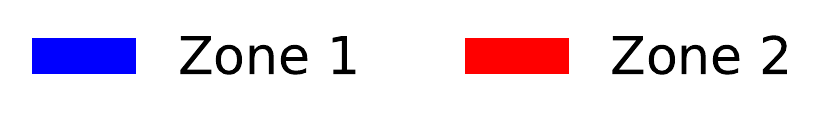}
   \end{adjustbox}
      \begin{adjustbox}{valign=t,minipage=0.47\textwidth}
         \hspace*{0.4cm}
         \hspace*{0.13cm}
         \includegraphics[width=0.9\textwidth]{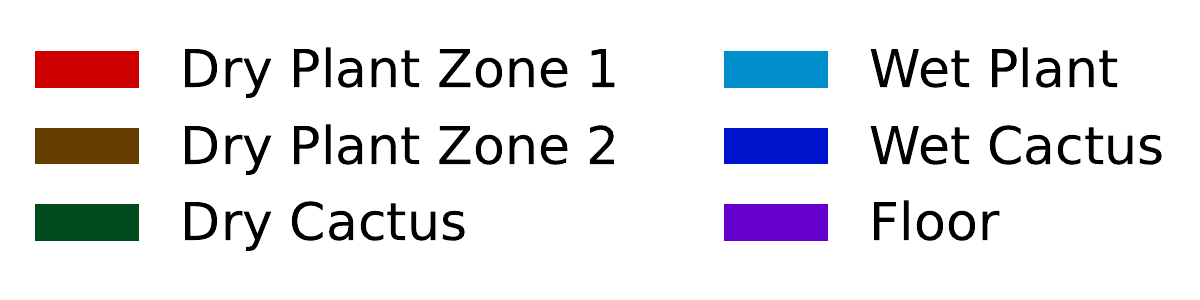}
   \end{adjustbox}
    \hfill
    \begin{adjustbox}{valign=t,minipage=0.47\textwidth}
         \vspace{-0.3cm}
         \includegraphics[width=\textwidth]{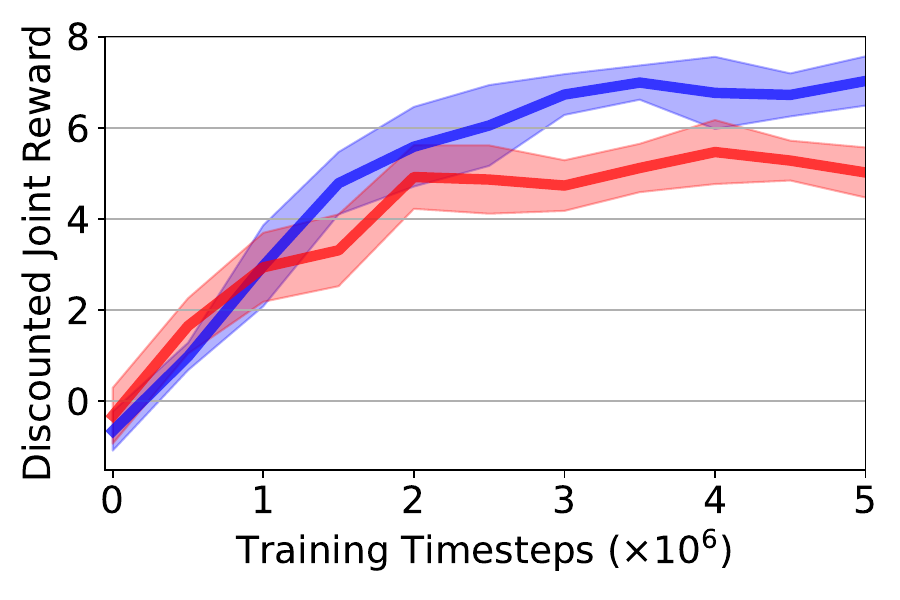}
         \subcaption{\footnotesize{Training rewards per zone in the half-room.}}
         \label{fig:half_room_training}
    \end{adjustbox}
    \begin{adjustbox}{valign=t,minipage=0.47\textwidth}
         \vspace{-0.3cm}
         \includegraphics[width=\textwidth]{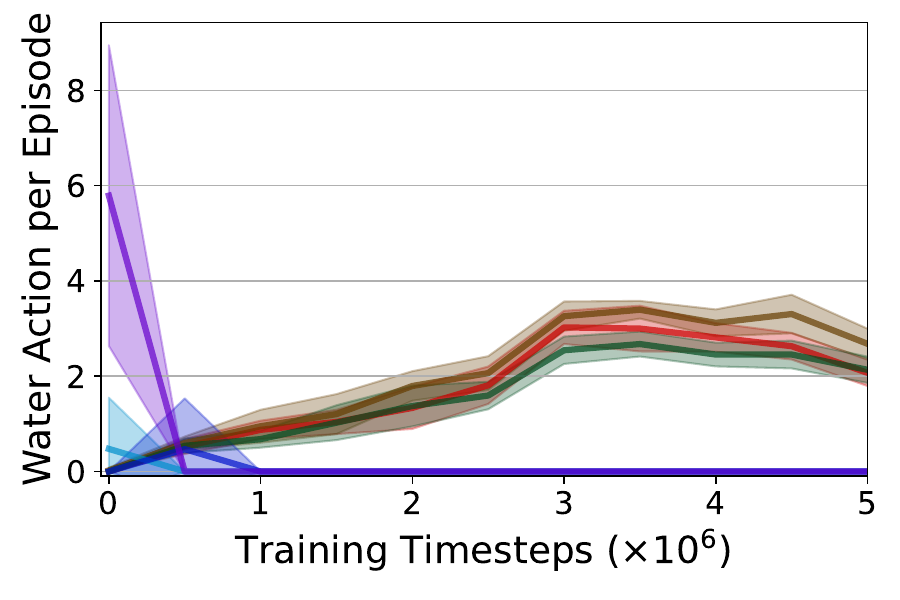}
         \subcaption{\footnotesize{Water action frequency in the plants \& cacti.}}
         \label{fig:cacti_room_frequency}
    \end{adjustbox}
        \caption{Results for the half-room and plants \& cacti environments. The bold line represents the mean over $10$ different seeds; the shaded area is a $95\%$ bootstrapped confidence interval.}
        \label{fig:combined_results}
        % \vspace{-0.7cm}
\end{figure*}

In this experiment, the monitor state is represented by a one-hot encoding $\{0, 1\}$, indicating whether a given state is monitored. The reward model is trained using environment states collected exclusively from Zone $1$.
% The reward model $\mathcal{R}(\EnvState, \EnvAction)$ is trained using samples in which the proxy reward is equal to the environment reward $\ProxyReward=\EnvReward$ (Zone 1).
To integrate monitor awareness into decision-making, the Q-network processes the environment state through two convolutional layers to extract features. These features are then concatenated with the one-hot encoding of the monitor state and passed through the last two fully connected layers, estimating the Q-value for each action.
By leveraging the neural network architecture and its capacity for generalization, the reward model generalizes from the monitored section (Zone $1$) to the unmonitored section (Zone $2$).
% This result is demonstrated by the agent's ability to appropriately water plants in both zones.
This generalization is evidenced by the agent's ability to appropriately water plants in both zones, despite only receiving reward signals from Zone 1 during training.
Since both zones contain an equal number of plants, comparing the training rewards reveals that rewards in both zones are comparable, indicating that the reward model not only learns in the monitored region but can also transfer its learned behavior to the unmonitored region, as shown in Figure~\ref{fig:half_room_training}. These results indicate effective generalization. However, a negligible difference in rewards exists between the two zones, which can be attributed to the presence of walls surrounding the room, given that some environment states in Zone $1$ appear visually different from those in Zone $2$.

This ability to generalize is particularly significant when considering the broader implications for Mon-MDPs. According to~\cite{parisi2024first}, the half-room environment is defined as an unsolvable Mon-MDP because the agent cannot observe rewards for nearly half of the environment states.
Thus, the agent cannot distinguish between two Mon-MDPs that have different optimal policies, \eg, whether plants should be watered on the right side of the room.
Therefore, we demonstrate that by combining a reward model with FA, the agent can infer unobserved rewards to learn a near-optimal policy. This highlights the potential of FA in overcoming partial reward observability, enabling agents to effectively navigate some unsolvable environments.

In the half-room environment, we excluded the $\bot=0$ and ``ignore'' approaches from Figure~\ref{fig:half_room_training} to keep the plots less cluttered and maintain focus on the generalization capabilities of the reward model. For completeness, a comparative figure including these approaches is provided in the appendix.
% enables generalization, resulting in a near-optimal policy. 
% \begin{wrapfigure}{r}{0.4\textwidth}
%     \begin{adjustbox}{valign=t,minipage=0.4\textwidth}
%         \centering
%          \includegraphics[width=\textwidth]{fig/half_room/legend_horiz.pdf}
%     \end{adjustbox}
%     \hfill
%     \begin{adjustbox}{valign=t,minipage=0.4\textwidth}
%         \centering
%         \includegraphics[width=\textwidth]{fig/half_room/training_reward_per_room_reward_cut_no_legend.pdf}
%     \end{adjustbox}
%         \caption{Training rewards divided by the zones in the half-room environment. The bold line represents the mean over $10$ seeds; the shaded area is $95\%$ interval confidence.}
%         \label{fig:half_room_training}
% \end{wrapfigure} 

\subsection{Plant-cactus environment}\label{subsec:plant_cactus_env}
This experiment explores a key limitation of FA for unsolvable Mon-MDPs: \emph{overgeneralization}, where agents incorrectly extrapolate rewards from monitored to unmonitored environment states, resulting in suboptimal or undesirable behavior, as states in Contribution~\ref{contribution:3}.
To study this phenomenon, we introduce a new plant to the plant-watering environment: \emph{cacti}. As illustrated in Figure~\ref{fig:cacti_room_env}, the monitored zone contains four plants, while the unmonitored zone includes four additional plants and four cacti (triangle pointing up). Unlike other plants, cacti have distinct watering requirements, and watering them incurs a penalty of $-1$. Notably, cacti have different representations ($[0, 1, 1]$) than standard plants ($[1, 1, 0]$) in how they are represented in the state tensor. However, the agent never observes any penalty for watering cacti because they are placed only in the unmonitored zone.

Figure~\ref{fig:cacti_room_frequency} presents water action frequency for plants, cacti, and the floor in the monitored and unmonitored zones throughout training. The y-axis represents the frequency of the watering action per episode. Initially, the agent spills water on the floor but quickly learns to avoid this due to the penalty. Similar to the half-room environment, the agent waters plants in both zones. However, despite their distinct state representation, it also waters cacti in the unmonitored zone at a similar rate.
This result demonstrates that FA can lead to incorrect reward extrapolation, causing the agent to adopt undesirable behaviors (\eg, watering cacti). This highlights the challenge outlined in Contribution~\ref{contribution:3} and underscores the need for strategies to mitigate overgeneralization in Mon-MDPs.

\subsection{Botanical garden environment}\label{subsec:botanical_garden_env}
Building on insights from the plant-cactus environment, where the reward model with FA exhibited a sub-optimal policy by watering cacti, the botanical garden environment explores Contribution~\ref{contribution:3}, aiming to evaluate whether agents can learn to act more cautiously in such settings. Specifically, exploring how agents treat novel plants when no reward is observable (in the unmonitored zone).
% whether agents can avoid watering novel plants in the unmonitored zone.

In the botanical garden environment, the monitored zone contains four standard plants and four cacti, allowing the agent to observe the outcomes of watering both. In contrast, the unmonitored zone includes four standard plants, four cacti, and four novel plants represented with a left-pointing triangle, as shown in Figure~\ref{fig:3_rooms_env}. The environment encodes plant types using three channels as follows: floor $[0, 0, 0]$, plants $[1, 1, 0]$, cacti $[0, 1, 1]$, and novel plants are sampled from $\{[0, 0, 1], [0, 1, 0], [1, 0, 0], [1, 0, 1], [1, 1, 1]\}$.

To explore the learning of cautious behavior, we adapt the learning to be cautious framework~\citep{mohammedalamen2021learning} to Mon-MDPs. First, an ensemble of $500$ reward models is trained on data from the monitored zone, with each reward model initialized randomly to capture reward uncertainty. Second, we optimize an approximation of CVaR using $k$-of-$N$ CFR~\citep{kofn},
where the $\nicefrac{k}{n}$ ratio determines the robustness level, ranging from risk-neutral $k = N$ to highly risk-averse $k = 1$ and $N \gg k$.
Our evaluation considers a range of policies, including that optimized from the original reward model, a mid-level robust policy $5$-of-$10$, and a highly robust policy $1$-of-$10$, focusing on the frequency of watering standard and novel plants across both zones.

Table~\ref{table:3_rooms_frequency_ratio} reports the mean and $95\%$ bootstrapped confidence interval (estimated by repeatedly resampling runs with replacement and computing the mean) of the watering frequency per episode for the reward model. Additionally, it reports the watering frequency ratio relative to the reward model for robust policies $5$-of-$10$ and $1$-of-$10$.
The reward model exhibits a higher watering frequency for standard plants in the unmonitored zone compared to the monitored zone. It also waters some novel plants $[0, 1, 0], [1, 0, 0]$, and $[1, 1, 1]$. Interestingly, the reward model waters $[1, 0, 0]$ more frequently than standard plants in the monitored zone, likely due to this representation being the inverse of cacti $[0, 1, 1]$, making it appear as a ``super-plant''. However, for other novel plants $[0, 0, 1]$ and $[1, 0, 1]$, the reward model never selects the watering action. This suggests that these plant representations inherently discourage watering.

In contrast, the robust policies $5$-of-$10$ and $1$-of-$10$, maintain a higher watering frequency for standard plants while significantly reducing watering for novel plants up to $5$ times less than the reward model. Moreover, increasing the degree of robustness leads to a greater reduction in watering novel plants. These results indicate that robust policies balance persisting in rewarding actions that are highly familiar while mitigating overgeneralization in novel situations. Consequently, the findings support Contribution~\ref{contribution:3}, demonstrating that learning a cautious policy mitigates overgeneralization.

% As shown in Table~\ref{table:3_rooms_frequency_ratio}, robust policies with lower $k$ values (indicating greater risk aversion) reduce the frequency of watering novel plants by up to $80\%$ compared to the risk-neutral policy on some novel plants. These results demonstrate that agents can learn to act cautiously in environments where rewards are unobservable, effectively mitigating the risks associated with overgeneralization. Therefore, demonstrating contribution $3$.\TODO{Monta: add a table \& explain more}

\begin{table}[tb]
  \centering
  \caption{Water action frequency per episode for the reward model and watering frequency ratio for robust policies $5$-of-$10$ and $1$-of-$10$ relative to the reward model, in the botanical garden environment. Represented by the mean and $95\%$ bootstrapping interval confidence over $30$ seeds. The last row shows novel plants that have never been watered by any policy.}
  \label{table:3_rooms_frequency_ratio}
  \begin{tabular}{|c|c|c|c|}
  \hline
    \textbf{Plant} & Reward model & Ratio $5$-of-$10$ &  Ratio $1$-of-$10$ \\ \hline
    Plants Zone $1$           & $3.28 \text{ } [3.00, 3.56]$ & $\times 1.30 \text{ } [1.30, 1.30]$ & $\times 1.29 \text{ } [1.29, 1.29]$ \\ \hline
    Plants Zone $2$           & $4.13 \text{ } [3.80, 4.43]$ & $\times 1.17 \text{ } [1.17, 1.17]$ & $\times 1.16 \text{ } [1.16, 1.16]$ \\ \hline
    $[0, 1, 0]$               & $0.01 \text{ } [0.00, 0.01]$ & $\times 0.27 \text{ } [0.27, 0.29]$ & $\times 0.25 \text{ } [0.25, 0.27]$ \\ \hline
    $[1, 0, 0]$               & $4.01 \text{ } [0.81, 8.02]$ & $\times 0.20 \text{ } [0.20, 0.23]$ & $\times 0.19 \text{ } [0.19, 0.20]$ \\ \hline
    $[1, 1, 1]$               & $0.10 \text{ } [0.06, 0.14]$ & $\times 0.56 \text{ } [0.56, 0.58]$ & $\times 0.41 \text{ } [0.41, 0.43]$ \\ \hline
    $[0, 0, 1]$, $[1, 0, 1]$  & $0.00 \text{ } [0.00, 0.00]$ & $\times 1.00 \text{ } [1.00, 1.00]$ & $\times 1.00 \text{ } [1.00, 1.00]$ \\ \hline
    \end{tabular}
    % \vspace{-0.7cm}
\end{table}

% \section{Limitation and Future Works}\label{sec:limitation}

\section{Conclusion and Future Work}\label{sec:conclusion}
Although the Mon-MDP framework aims to model real-world applications where traditional MDPs fall short due to unobservable rewards, prior work has been limited to tabular environments with a finite, enumerable number of states.
However, most real-world applications do not have finite state or action spaces or have an enormous number of states and actions. This work explores the challenges of non-tabular representations in Mon-MDPs and investigates function approximation capabilities and limitations.  Our results demonstrate that training a reward model with FA allows agents to achieve near-optimal policies in some environments previously labeled as unsolvable. However, we also identify a critical limitation of FA: \emph{overgeneralization}, where the model incorrectly extrapolates rewards, leading to suboptimal and potentially unsafe behaviors. To mitigate this, we adopt a robust policy optimization approach leveraging reward uncertainty and $k$-of-$N$ CFR, demonstrating that agents can learn to act cautiously, even when rewards are unobservable.

Moving forward, our research paves the way for multiple directions. First, extending Mon-MDPs to policy-based and actor-critic methods to handle continuous action spaces, including REINFORCE, TRPO, PPO, and SAC~\citep{sutton1999policy, konda1999actor,schulman15trpo, schulman2017proximal, haarnoja2018soft}. Second, adopting computationally efficient approaches for capturing epistemic uncertainty, \eg, noisy networks~\parencite{fortunato2018noisy} and epistemic neural networks~\parencite{osband2023epistemic}. Third, investigating the phenomenon of plasticity loss in Mon-MDPs when using deep neural networks --- as we observed hints of that possibility in our experiments, see Appendix~\ref{app:subsec:plasticity} --- and exploring whether existing mitigation strategies in traditional MDPs~\citep{abbas2023loss,nikishin2023deep,dohare2024loss,elsayed2024addressing} can be effectively adapted to Mon-MDPs. Finally, validating Mon-MDPs in real-world applications, such as industrial automation and autonomous robotics, to assess their practical feasibility and effectiveness.
% will be crucial in assessing their practical impact and scalability.
By addressing these challenges, we move toward making Mon-MDPs a more robust framework for decision-making in complex and realistic environments with partially observable rewards.

\section{Broader Impact Statement}\label{sec:broaderImpact}
This work seeks to advance the field of RL by enabling the deployment of safe, reliable, and robust AI systems capable of navigating real-world complexities. Furthermore, this work sets the foundation for future research in settings where rewards are not always present.
However, the community should be careful \emph{not to solely rely} on such methods, as the automated safety measures are meant to complement, \emph{not replace}, human judgment and safety planning.

% \subsubsection*{Broader Impact Statement}
% \label{sec:broaderImpact}
% This work seeks to advance the field of RL by enabling the deployment of safe, reliable, and robust AI systems capable of navigating real-world complexities. Furthermore, this work sets the foundation for future research in settings where rewards are not always present.
% % We believe this work primarily helps us build safer and more reliable systems.
% However, the community should be careful not to solely rely on such methods.

% \begin{ack}\label{sec:ack}
% Computation provided by Digital Research Alliance of Canada.
% Thanks to Matthew E. Taylor and John D. Martin for constructive feedback on an early draft of this paper.
% \end{ack}

% \section*{References}
{
\small
\bibliography{references}
}

%%%%%%%%%%%%%%%%%%%%%%%%%%%%%%%%%%%%%%%%%%%%%%%%%%%%%%%%%%%%
\newpage
\beginSupplementaryMaterials
\appendix
\section{Experiments}\label{app:experiments}
Our neural network architecture, used for the reward model, $\bot=0$, and ``ignore'' methods, consists of two convolutional layers followed by two fully connected layers, with rectified linear unit (ReLU) activations applied between them~\cite{glorot2011deep}. The complete network architecture is detailed in Table~\ref{table:network_arch}, and hyper-parameters for all our experiments are summarized in Table~\ref{table:parameters_values}.

The networks are trained using the Adam optimizer~\cite{kingma2014adam} to minimize the mean squared error (MSE), where the reward model aims to minimize the error between predicted and actual rewards, while Q-models minimize the error between estimated and actual discounted returns.

Our code is provided as a ZIP file included in the supplementary materials. The \texttt{code} directory contains scripts for training reward models, Q-networks, and ensembles of reward models, as well as for performing $k$of-$N$ CFR optimization using PyTorch~\cite{paszke2017automatic}.
% \href{https://drive.google.com/drive/folders/10VL5ZZuPO6RlzxAJUXpEiS2oyTuUSu8b}{link}\footnote{\url{https://shorturl.at/aQF9V}}:
% \begin{itemize}
%     \item The \texttt{code} contains scripts for training reward models, Q-networks, and ensembles of reward models, and $k$of-$N$ CFR optimization using PyTorch~\cite{paszke2017automatic}.
%     \item The \texttt{models} directory includes pre-trained reward models and Q-networks, with models for each experiment stored in separate subdirectories.
% \end{itemize}

All our agents and models were trained on a $2.65$GHz AMD\textregistered{} EPYC\texttrademark{} 7413 (Zen $2$) CPU with $50$ GB of memory, with access to $\frac{1}{5}$ of an Nvidia\textregistered{} A100SXM4\texttrademark{} GPU.
Specifically, agents for the Binary, Half-room, and Plant-cactus environments were trained over a duration of $3$ hours.
While the Botanical garden environment, an ensemble of $500$ reward models was trained, each requiring approximately $10$ minutes, resulting in a total of $83.3$ GPU hours.
Additionally, $50$ iterations of the $k$-of-$N$ CFR algorithm were executed on the CPU in under $15$ seconds.

\begin{table}[ht]
  \centering
  \caption{Hyper-parameters values}
  \label{table:parameters_values}
\begin{tabular}{lc}
  \toprule
  Hyper-parameter    &    Value   \\\midrule
  Discount factor $\gamma$ & 0.99\\
  Start training after   &    $10^5$ timesteps     \\
  Replay buffer size  &    $10^6$ samples   \\
  Target network update frequency   &     $50$  \\
  Batch size & $512$ samples\\
  Q-network update frequency & $250$ episodes \\
  Reward model learning rate ($\eta^{R}$) & $10^{-4}$\\
  Q-network learning rate ($\eta^{Q}$) & $10^{-4}$\\
  Initial exploration rate $\epsilon$ & $1.0$\\
  Minimum exploration rate $\epsilon$ & $0.1$\\
  Exploration $\epsilon$ linear decay rate & $10^{-4}$\\
  Weight decay & $0.0$\\
  Optimizer & Adam~\cite{kingma2014adam} \\  
  \bottomrule
\end{tabular}
\end{table}

\begin{table}[ht]
  \centering
  \caption{Network architecture}
  \label{table:network_arch}
\begin{tabular}{lc}
  \toprule
  First convolution layer kernel size & $5$\\
  First convolution layer stride & $1$\\
  First convolution layer number of channels & $32$\\
  Second convolution layer kernel size & $3$\\
  Second convolution layer stride & $1$\\
  Second convolution layer number of channels & $64$\\
  Number of neurons in the hidden layer & $512$ \\  
  \bottomrule
\end{tabular}
\end{table}

\subsection{Hyper-parameters tuning}\label{app:subsec:parameters_parameters_tuning}

In addition to the reward model, we evaluate two alternative methods that treat Mon-MDPs as traditional MDPs: 
\begin{enumerate}
    \item $\bot=0$, undefined rewards are replaced with $0$ when updating the Q-network.
    \item ``Ignore'', updates the Q-network only with samples where rewards are observable, discarding samples with unobservable rewards.
\end{enumerate}

For a fair comparison, all algorithms employ the same DQN architecture as the reward model. The learning rate for both the reward model and Q-network, denoted as $\eta^{R}, \eta^{Q}$, are set to $0.0001$.
Additional experiments are conducted using alternative learning rates of ($0.0005$, $0.001$, and $0.005$). The area under the training curve for each setting is summarized in Table~\ref{table:binary_auc_parameters}, with the mean and a $95\%$ interval confidence reported.

\begin{figure*}[ht]
  \centering
  \begin{adjustbox}{valign=t,minipage=0.9\textwidth}
         \includegraphics[width=\textwidth]{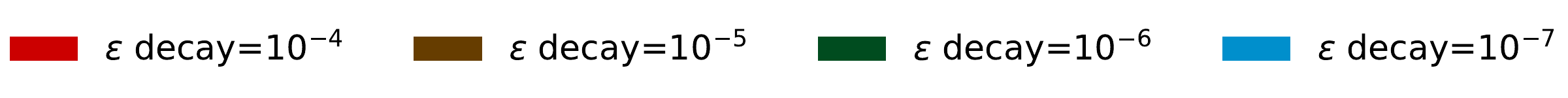}
    \end{adjustbox}
  \hfill
  \begin{adjustbox}{valign=t,minipage=0.47\textwidth}
         \vspace{-0.2cm}
         \includegraphics[width=\textwidth]{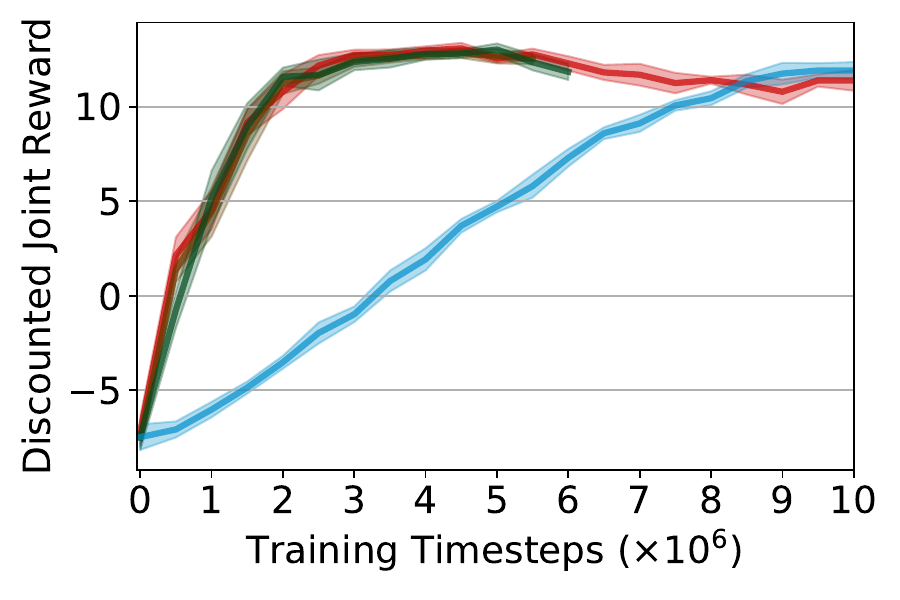}
         \subcaption{\footnotesize{Training curve for different exploration $\epsilon$ decay rates in the binary environment.}}
         \label{fig:appendix_binary_eps}
    \end{adjustbox}
    \begin{adjustbox}{valign=t,minipage=0.47\textwidth}
         \vspace{-0.2cm}
         \includegraphics[width=\textwidth]{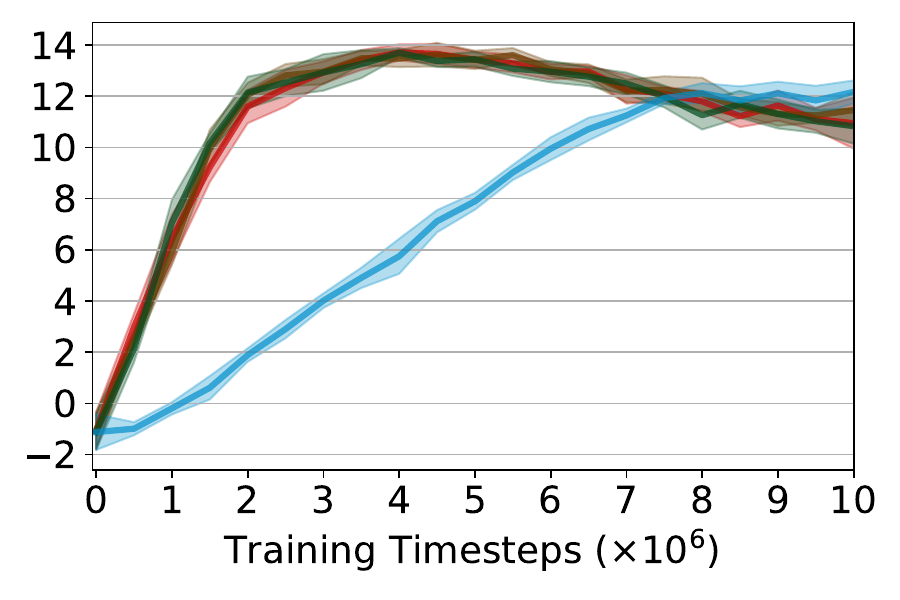}
         \subcaption{\footnotesize{Training curve for different exploration $\epsilon$ decay rates in the half-room environment.}}
         \label{fig:appendix_room_eps}
    \end{adjustbox}
        \caption{Exploration $\epsilon$ decay rate tuning in the binary and half-room environments. The bold line represents the mean over $10$ seeds; the shaded area is a $95\%$ interval confidence.}
        \label{fig:appendix_binary_room_eps}
\end{figure*}

Exploration follows an $\epsilon$-greedy strategy, where $\epsilon$ starts at $1.0$ and linearly decays to $0.1$ over time.
In the binary environment, we tune the decay rate as illustrated in Figure~\ref{fig:appendix_binary_eps}.
Similarly, Figure~\ref{fig:appendix_room_eps} illustrates the tuning of the exploration decay rate for the half-room environment.
Both figures compare different exploration decay rates ${10^{-4}, 5*10^{-4}, 10^{-3}, 5*10^{-3}}$.
In both environments, smaller exploration decay rates lead to a lower area under due to an increased number of exploratory steps.
However, this additional exploration does not provide significant benefits, as both environments are relatively simple to explore.

\begin{table}[ht]
  \centering
  \caption{Area under the training curve for the reward model: hyper-parameters tuning in the binary environment, reporting the mean and a $95\%$ interval confidence.}
  \label{table:binary_auc_parameters}
  \begin{tabular}{|c|c|c|}
  \hline
  $\eta^{Q}$  & $\eta^{R}$ & Area under the training curve \\[2ex] \hline
  $10^{-4}$   & $10^{-4}$   & $20.00 \text{ } [19.81, 20.18]$ \\
  $10^{-4}$   & $5*10^{-4}$ & $19.79 \text{ } [19.62, 19.96]$ \\
  $10^{-4}$   & $10^{-3}$   & $19.69 \text{ } [19.32, 20.04]$ \\
  $10^{-4}$   & $5*10^{-3}$ & $19.58 \text{ } [19.34, 19.82]$ \\ \hline
  $5*10^{-4}$ & $10^{-4}$   & $19.35 \text{ } [19.18, 19.54]$ \\
  $5*10^{-4}$ & $5*10^{-4}$ & $19.54 \text{ } [19.34, 19.76]$ \\
  $5*10^{-4}$ & $10^{-3}$   & $19.27 \text{ } [19.03, 19.51]$ \\
  $5*10^{-4}$ & $5*10^{-3}$ & $19.20 \text{ } [18.94, 19.48]$ \\ \hline
  $10^{-3}$   & $10^{-4}$   & $18.69 \text{ } [18.38, 18.95]$ \\
  $10^{-3}$   & $5*10^{-4}$ & $18.92 \text{ } [18.64, 19.14]$ \\
  $10^{-3}$   & $10^{-3}$   & $18.45 \text{ } [18.16, 18.80]$ \\
  $10^{-3}$   & $5*10^{-3}$ & $17.62 \text{ } [14.76, 19.17]$ \\ \hline
  $5*10^{-3}$ & $10^{-4}$   & $14.04 \text{ } [10.94, 15.82]$ \\
  $5*10^{-3}$ & $5*10^{-4}$ & $15.36 \text{ } [14.63, 16.14]$ \\
  $5*10^{-3}$ & $10^{-3}$   & $11.25 \text{ } [6.58, 14.75]$ \\
  $5*10^{-3}$ & $5*10^{-3}$ & $5.51 \text{ } [1.11, 9.96]$ \\ \hline
    \end{tabular}
\end{table}

\subsection[Reward model, ``ignore'' and zero performance]{Reward model, ``ignore'' and $\bot=0$ performance}\label{app:subsuec:baselines}
\begin{figure*}[ht]
  \centering
  \begin{adjustbox}{valign=t,minipage=0.65\textwidth}
         \includegraphics[width=\textwidth]{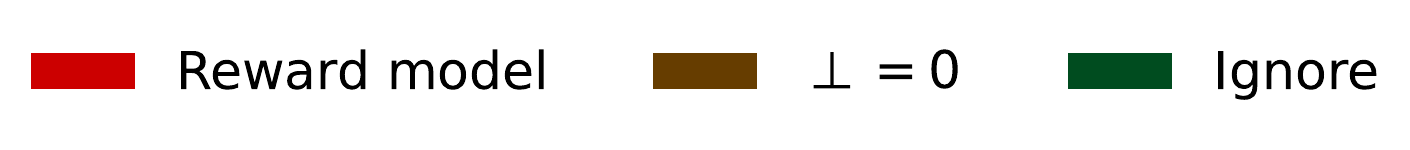}
    \end{adjustbox}
  \hfill
  \begin{adjustbox}{valign=t,minipage=0.47\textwidth}
         \vspace{-0.35cm}
         \includegraphics[width=\textwidth]{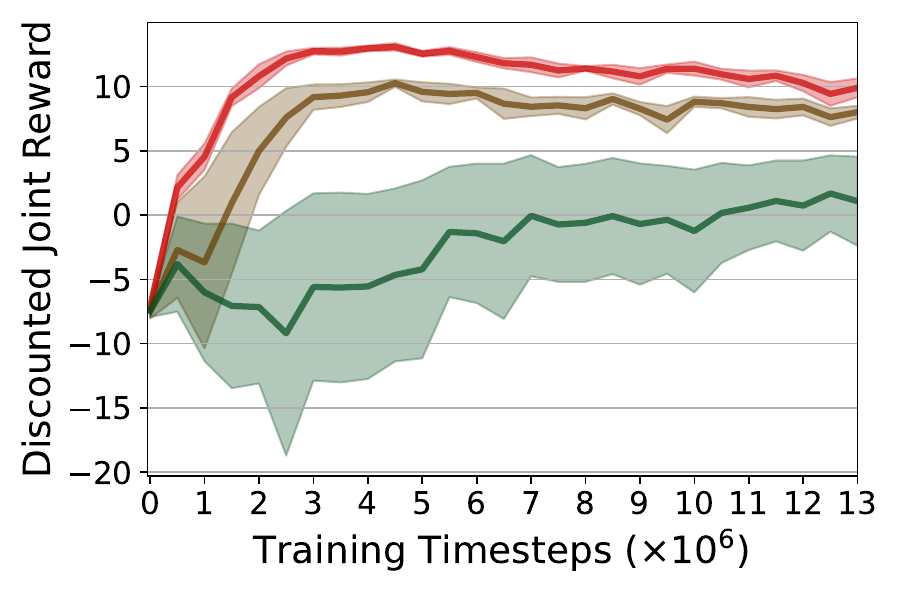}
         \subcaption{\footnotesize{Training curve in the binary environment.}}
         \label{fig:appendix_binary_train_conf}
    \end{adjustbox}
    \begin{adjustbox}{valign=t,minipage=0.47\textwidth}
         \vspace{-0.35cm}
         \includegraphics[width=\textwidth]{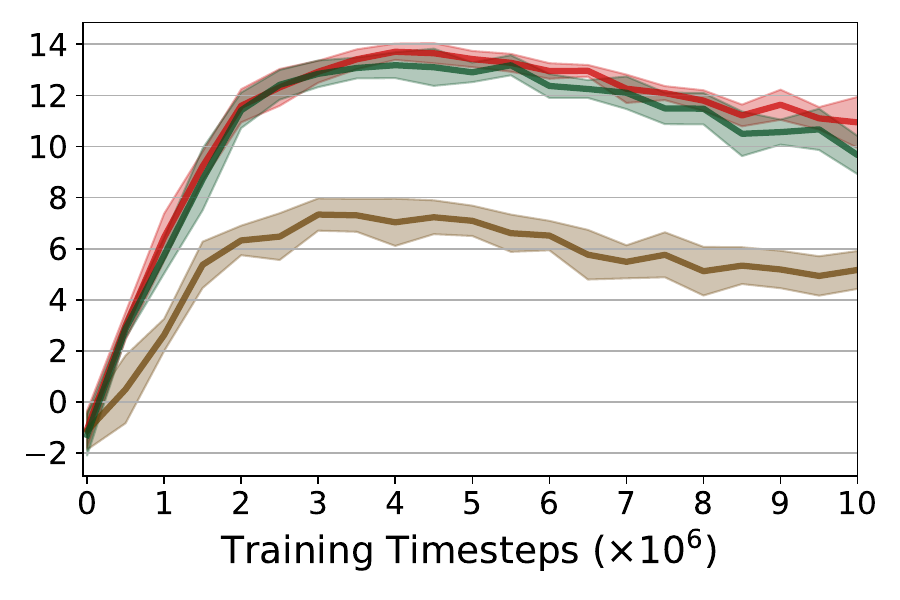}
         \subcaption{\footnotesize{Training curve in the half-room environment.}}
         \label{fig:appendix_room_train_conf}
    \end{adjustbox}
        \caption{Reward model, ``ignore'', and $\bot=0$ training curves in the binary and half-room environments. The bold line represents the mean over $10$ seeds; the shaded area is a $95\%$ interval confidence.}
        \label{fig:appendix_binary_room_train_conf}
\end{figure*}

\begin{figure*}[ht]
  \centering
  % \vspace{-0.7cm}
  \begin{adjustbox}{valign=t,minipage=0.65\textwidth}
         \includegraphics[width=\textwidth]{fig/appendix/baselines/baselines_legend.pdf}
    \end{adjustbox}
  \hfill
  \begin{adjustbox}{valign=t,minipage=0.47\textwidth}
         \vspace{-0.35cm}
         \includegraphics[width=\textwidth]{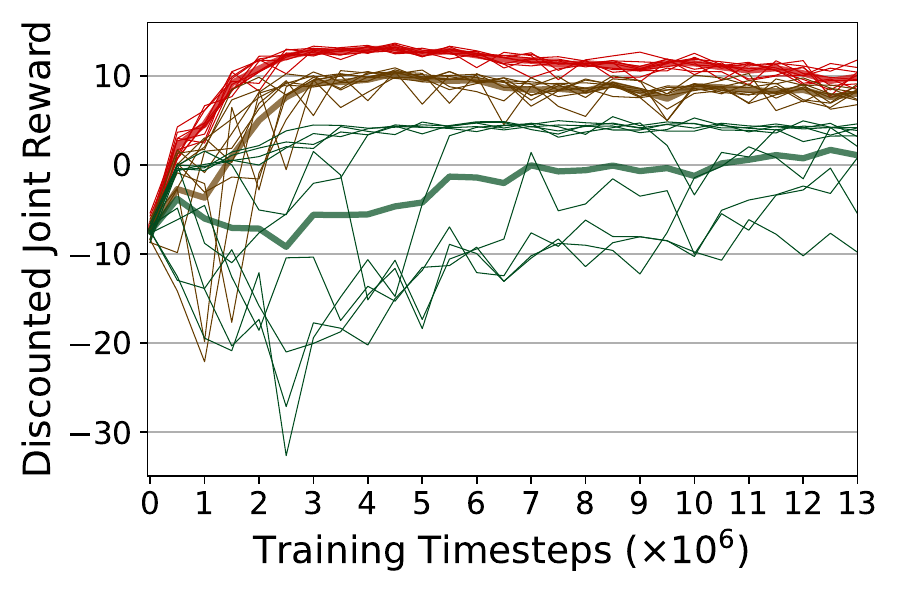}
         \subcaption{\footnotesize{Training curve in the binary environment.}}
         \label{fig:appendix_binary_train_indv_run}
    \end{adjustbox}
    \begin{adjustbox}{valign=t,minipage=0.47\textwidth}
         \vspace{-0.35cm}
         \includegraphics[width=\textwidth]{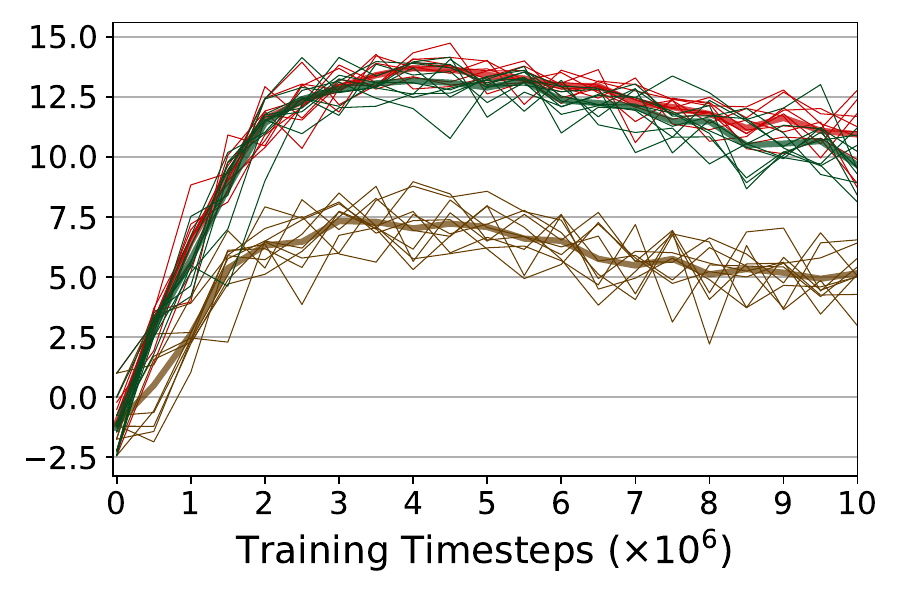}
         \subcaption{\footnotesize{Training curve in the half-room environment.}}
         \label{fig:appendix_room_train_indv_run}
    \end{adjustbox}
        \caption{Reward model, ``ignore'', and $\bot=0$ training curves in the binary and half-room environments. The bold line represents the mean over $10$ seeds, and each line represents a seed.}
        \label{fig:appendix_binary_room_train_indv_run}
\end{figure*}

In the binary environment, in addition to the reward model, we evaluate $\bot=0$, replacing undefined rewards with $0$, and ``ignore'' updating the Q-network only with samples where rewards are observable.
For a fair comparison, all algorithms utilize the same DQN architecture as the reward model, and performance is reported as the mean with a $95\%$ confidence interval. This is illustrated in Figure~\ref{fig:appendix_binary_room_train_conf}, where the $95\%$ confidence interval is represented as a shaded area, and in Figure~\ref{fig:appendix_binary_room_train_indv_run}, which shows the individual runs.

\subsection{Detailed results for the Botanical garden environment}\label{app:subsec:botanical_garden_env}
The Botanical Garden environment serves as a benchmark for evaluating whether agents can learn to act more cautiously in such settings. Specifically, exploring how agents treat novel plants when no reward is observable (in the unmonitored zone).

Table~\ref{table:3_rooms_frequency} reports the average watering action frequency per episode for bother standard and novel plants, reporting the mean watering frequency per episode, along with 95\% bootstrapping confidence intervals, over $30$ random seeds, for different approaches:
\begin{enumerate}
    \item The reward model-based policy relies solely on learned reward estimates.
    \item The $10$-of-$10$ policy considers the expected (average) outcomes across all sampled reward models, leading to a risk-neutral policy.
    \item The $5$-of-$10$ policy considers the worst $50\%$ of the reward models, leading to a mid-risk-averse policy.
    \item The $1$-of-$10$ policy considers the worst $10\%$ of the reward models, leading to a highly risk-averse policy.
\end{enumerate}

In both zones, robust policies consistently increase watering frequency compared to the reward model. However, as the degree of robustness increases (more risk-averse), watering frequency decreases for standard plants in both zones.
For novel plants in Zone $2$ (unmonitored zone), robust policies further reduce watering frequency, with this effect becoming more evident as robustness increases.
Interestingly, even the risk-neutral $10$-of-$10$ policy, which optimizes for the expected distribution, exhibits some caution when dealing with novel plants. It waters them less frequently than the reward model, potentially leveraging epistemic uncertainty across all reward models.
Notably, for certain novel plant states like $[0, 0, 1]$ and $[1, 0, 1]$, none of the policies apply watering actions, suggesting that these plant representations inherently discourage watering.

Additionally, Table~\ref{table:3_rooms_frequency_mean_ratio_full} presents the same results in a different format, showing the ratio of watering actions for robust policies $10$-of-$10$, $5$-of-$10$, and $1$-of-$10$ relative to the reward model. This comparison highlights how each policy varies from the reward model in terms of watering frequency reduction.

\begin{table}[ht]
  \centering
  \caption{Water action frequency per episode for the reward model and robust policies $10$-of-$10$, $5$-of-$10$, and $1$-of-$10$, in the botanical garden environment. Represented by the mean and $95\%$ bootstrapped interval confidence over $30$ seeds. The last row shows novel plants that have never been watered by any policy.}
  \label{table:3_rooms_frequency}
  \begin{tabular}{|c|c|c|c|c|}
  \hline
    \textbf{Plant} & Reward model & $10$-of-$10$ & $5$-of-$10$ & $1$-of-$10$ \\ \hline
    Plants Zone $1$           & $3.28 \text{ } [3.00, 3.56]$ & $4.78 \text{ } [4.44, 5.09]$ & $4.27 \text{ } [3.85, 4.68]$ & $4.22 \text{ } [3.83, 4.55]$ \\ \hline
    Plants Zone $2$           & $4.13 \text{ } [3.80, 4.43]$ & $5.45 \text{ } [5.08, 5.78]$ & $4.83 \text{ } [4.31, 5.24]$ & $4.77 \text{ } [4.41, 5.14]$ \\ \hline
    $[0, 1, 0]$               & $0.01 \text{ } [0.00, 0.01]$ & $0.00 \text{ } [0.00, 0.00]$ & $0.00 \text{ } [0.00, 0.00]$ & $0.00 \text{ } [0.00, 0.00]$ \\ \hline
    $[1, 0, 0]$               & $4.01 \text{ } [0.81, 8.02]$ & $1.23 \text{ } [1.10, 1.37]$ & $0.82 \text{ } [0.73, 0.90]$ & $0.75 \text{ } [0.69, 0.80]$ \\ \hline
    $[1, 1, 1]$               & $0.10 \text{ } [0.06, 0.14]$ & $0.10 \text{ } [0.09, 0.11]$ & $0.05 \text{ } [0.05, 0.06]$ & $0.04 \text{ } [0.03, 0.05]$ \\ \hline
    $[0, 0, 1]$, $[1, 0, 1]$  & $0.00 \text{ } [0.00, 0.00]$ & $0.00 \text{ } [0.00, 0.00]$ & $0.00 \text{ } [0.00, 0.00]$ & $0.00 \text{ } [0.00, 0.00]$ \\ \hline
  \end{tabular}
\end{table}

\begin{table}[ht]
  \centering
  \caption{Water action frequency per episode for the reward model and the ratio for robust policies $10$-of-$10$, $5$-of-$10$, and $1$-of-$10$ w.r.t the reward model, in the botanical garden environment. Represented by the mean and $95\%$ bootstrapped interval confidence over $30$ seeds. The last row shows novel plants that have never been watered by any policy.}
  \label{table:3_rooms_frequency_mean_ratio_full}
  \begin{tabular}{|c|c|c|c|c|}
  \hline
    \textbf{Plant} & Reward model & Ratio $10$-of-$10$ & Ratio $5$-of-$10$ & Ratio $1$-of-$10$ \\ \hline
    Plants Zone $1$           & $3.28 \text{ } [3.00, 3.56]$ & $\times 1.46 \text{ } [1.46, 1.46]$ & $\times 1.30 \text{ } [1.30, 1.30]$ & $\times 1.29 \text{ } [1.29, 1.29]$ \\ \hline
    Plants Zone $2$           & $4.13 \text{ } [3.80, 4.43]$ & $\times 1.32 \text{ } [1.32, 1.32]$ & $\times 1.17 \text{ } [1.17, 1.17]$ & $\times 1.16 \text{ } [1.16, 1.16]$ \\ \hline
    $[0, 1, 0]$               & $0.01 \text{ } [0.00, 0.01]$ & $\times 0.40 \text{ } [0.40, 0.41]$ & $\times 0.27 \text{ } [0.27, 0.29]$ & $\times 0.25 \text{ } [0.25, 0.27]$ \\ \hline
    $[1, 0, 0]$               & $4.01 \text{ } [0.81, 8.02]$ & $\times 0.31 \text{ } [0.31, 0.34]$ & $\times 0.20 \text{ } [0.20, 0.23]$ & $\times 0.19 \text{ } [0.19, 0.20]$ \\ \hline
    $[1, 1, 1]$               & $0.10 \text{ } [0.06, 0.14]$ & $\times 0.99 \text{ } [0.99, 1.04]$ & $\times 0.56 \text{ } [0.56, 0.58]$ & $\times 0.41 \text{ } [0.41, 0.43]$ \\ \hline
    $[0, 0, 1]$, $[1, 0, 1]$  & $0.00 \text{ } [0.00, 0.00]$ & $\times 1.00 \text{ } [1.00, 1.00]$ & $\times 1.00 \text{ } [1.00, 1.00]$ & $\times 1.00 \text{ } [1.00, 1.00]$ \\ \hline
  \end{tabular}
\end{table}

\subsection{Plasticity loss in Mon-MDPs}\label{app:subsec:plasticity}

Plasticity loss refers to the loss of flexibility in an agent's learning process as it encounters more samples. This phenomenon occurs when the agent becomes too committed to previously learned behaviors or strategies, preventing it from adapting effectively to new, unforeseen states or changes in the environment.

In Mon-MDPs, we hypothesize that plasticity loss is particularly noticeable because the agent learns two models: i) a reward model to overcome unobservable rewards, which influences the Q-network by providing reward predictions, and ii) the Q-network itself, which updates its values based on both state and reward inputs.

\begin{figure*}[ht]
  \begin{adjustbox}{valign=t,minipage=0.47\textwidth}
         \includegraphics[width=\textwidth]{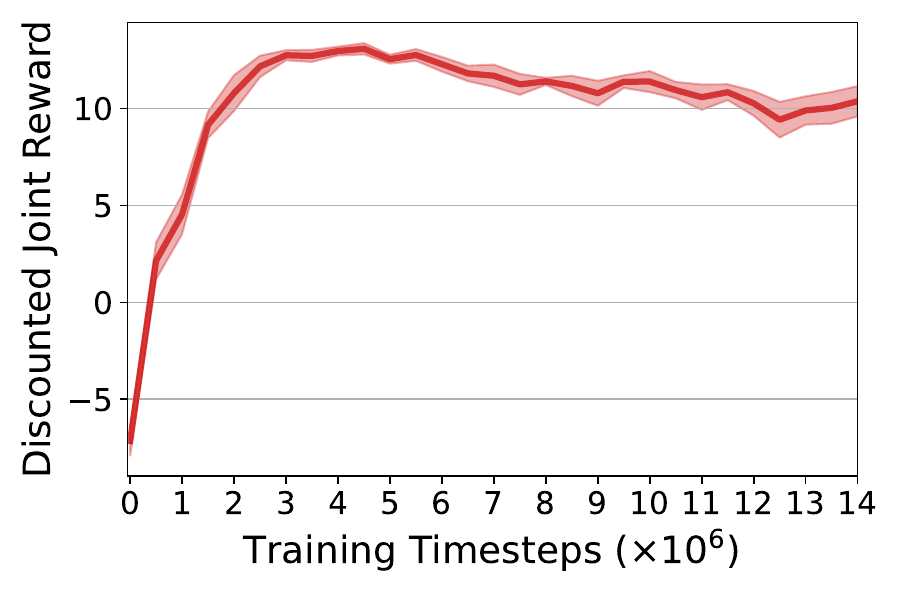}
         \subcaption{\footnotesize{Training curve.}}
         \label{fig:appendix_plasticity_binary_train}
    \end{adjustbox}
    \begin{adjustbox}{valign=t,minipage=0.47\textwidth}
         \includegraphics[width=\textwidth]{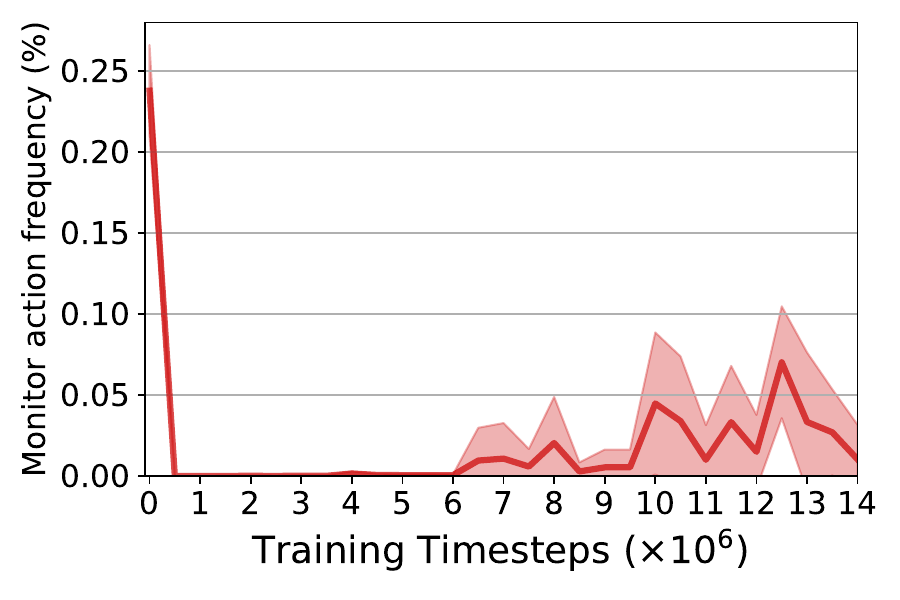}
         \subcaption{\footnotesize{Frequency of monitor action during training.}}
         \label{fig:appendix_plasticity_binary_frequency}
    \end{adjustbox}
        \caption{Preliminary evidence for the loss of plasticity in the binary environment. The bold line represents the mean over $10$ seeds; the shaded area is a $95\%$ interval confidence.}
        \label{fig:appendix_plasticity_binary}
\end{figure*}

\begin{figure*}[ht]
\centering
    \begin{adjustbox}{valign=t,minipage=0.6\textwidth}
        \hspace{1.2cm}
         \includegraphics[width=0.7\textwidth]{fig/half_room/legend_horiz.pdf}
   \end{adjustbox}
   \vfill
  \begin{adjustbox}{valign=t,minipage=0.6\textwidth}
        \vspace{-0.3cm}
        \centering
         \includegraphics[width=\textwidth]{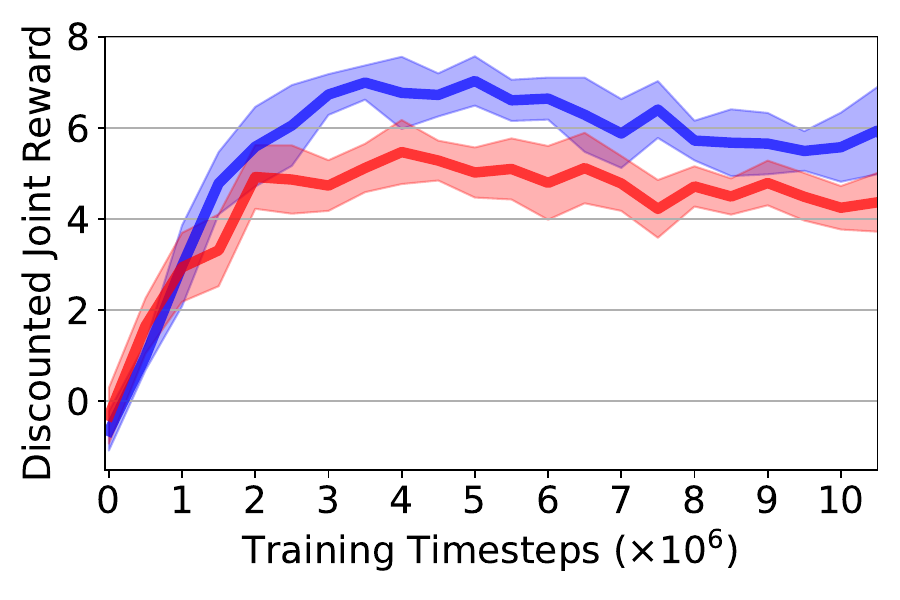}
    \end{adjustbox}
        \caption{Preliminary evidence for the loss of plasticity in the half-room environment. The bold line represents the mean over $10$ seeds; the shaded area is a $95\%$ interval confidence.}
        \label{fig:appendix_plasticity_half_room}
\end{figure*}

Our experiments demonstrate this effect. In the binary environment, after $6$M timesteps, the agent’s performance deteriorates, as indicated by the drop in training reward in Figure~\ref{fig:appendix_plasticity_binary}. The agent even starts requesting to be monitored, further confirming its declining adaptability. Similarly, in the half-room environment, training beyond $5$M-$7$M timesteps leads to a decline in accumulated rewards across both zones, as shown in Figure~\ref{fig:appendix_plasticity_half_room}.

These findings suggest that the use of deep learning in Mon-MDPs may contribute to plasticity loss. As future work, we plan to investigate this phenomenon further and explore whether existing mitigation strategies from traditional MDPs~\citep{abbas2023loss,nikishin2023deep,dohare2024loss,elsayed2024addressing} can be effectively adapted to Mon-MDPs.

\end{document}